\documentclass[11pt]{article}

\usepackage[margin=1in]{geometry}

\usepackage{enumerate}
\usepackage[autosize]{dot2texi}
\usepackage{tikz}
\usetikzlibrary{shapes,arrows}
\usepackage{amsthm,amsfonts,mathtools}
\usepackage{amsmath}
\usepackage{amssymb}
\usepackage{latexsym}
\usepackage{verbatim}
\usepackage{xcolor}
\usepackage{booktabs}

\usepackage{multirow}
\usepackage{mathrsfs}
\usepackage{graphicx}
\usepackage{subfigure}
\usepackage{hyperref}
\usepackage{bbm}
\usepackage{dsfont}
\usepackage{tipa,bm}
\usepackage[linesnumbered,vlined,ruled]{algorithm2e}
\usepackage{textcomp}
\usepackage{soul}

\usepackage[numbers]{natbib}

\DeclareMathOperator{\MVN}{\mathcal{MVN}}
\DeclareMathOperator{\cov}{\text{cov}}
\DeclareMathOperator{\var}{\text{var}}

\DeclareMathOperator{\BIC}{\text{BIC}}

\DeclareMathOperator{\BF}{\text{BF}}

\DeclareMathOperator{\RBF}{\text{RBF}}
\DeclareMathOperator{\Lin}{\text{Lin}}

\newcommand{\AR}{\text{AR}}
\DeclareMathOperator{\Min}{\text{Min}}
\DeclareMathOperator{\Meh}{\text{Meh}}
\DeclareMathOperator{\Chy}{\text{Chy}}

\newcommand{\M}{\text{M}}
\newcommand{\R}{\mathbb{R}}
\newcommand{\N}{\mathbb{N}}
\newcommand{\E}{\mathbb{E}}
\newcommand{\bx}{\mathbf{x}}
\newcommand{\eps}{\epsilon}

\newcommand{\bX}{\mathbf{X}}

\newcommand{\cD}{\mathcal{D}}
\newcommand{\cDrob}{\mathcal{D}_{\textrm{rob}}}
\newcommand{\cK}{\mathcal{K}}

\newcommand{\GP}{\mathcal{G}\mathcal{P}}

\newtheorem{rem}{Remark}

\definecolor{blue1}{RGB}{205,240,255}
\definecolor{blue2}{RGB}{150,210,255}
\definecolor{yellow1}{RGB}{255,253,210}
\definecolor{yellow2}{RGB}{255,250,190}
\definecolor{red1}{RGB}{255,220,220}
\definecolor{red2}{RGB}{255,200,200}
\definecolor{red3}{RGB}{255,50,50}
\definecolor{red4}{RGB}{0,0,0}
\definecolor{white}{RGB}{255, 255, 255}

\title{Expressive Mortality Models through Gaussian Process Kernels}

\author{
  Jimmy Risk \\
  Mathematics \& Statistics \\
  Cal Poly Pomona \\
  Pomona CA 91676\\
  \texttt{jrisk@cpp.edu} \\
  \and  Mike Ludkovski \\
  Statistics \& Applied Probability \\
  University of California \\
  Santa Barbara, CA 93106-3110\\
  \texttt{ludkovski@pstat.ucsb.edu}
}

\begin{document}

\maketitle

\begin{abstract} 
We develop a flexible Gaussian Process (GP) framework for learning the covariance structure of Age- and Year-specific mortality surfaces. Utilizing the additive and multiplicative structure of GP kernels, we design a genetic programming algorithm to search for the most expressive kernel for a given population. Our compositional search builds off the Age-Period-Cohort (APC) paradigm to construct a covariance prior best matching the spatio-temporal dynamics of a mortality dataset. We apply the resulting genetic algorithm (GA) on synthetic case studies to validate the ability of the GA to recover APC structure, and on real-life national-level datasets from the Human Mortality Database. Our machine-learning based analysis provides novel insight into the presence/absence of Cohort effects in different populations, and into the relative smoothness of mortality surfaces along the Age and Year dimensions. Our modelling work is done with the PyTorch libraries in Python and provides an in-depth investigation of employing GA to aid in compositional kernel search for GP surrogates.
\end{abstract}


\section{Introduction}

Modeling population mortality rates as a function of Age and Calendar time has been an area of active research for over a century. In 1992, the Lee-Carter model \cite{lee2000lee} introduced a stochastic \emph{Period effect} for the temporal evolution of mortality, spurring an explosion of statistical and probabilistic approaches to this topic. The Age-Period structure proved insufficient for many applications, leading to the development of Cohort-based extensions \cite{renshaw2006cohort} whereby mortality is further differentiated by birth year. These Age-Period-Cohort (APC) models \cite{hunt2020identifiability,cairns2011mortality} have become an essential tool for analyzing mortality trends while disentangling the effects of the respective three main predictors.

Gaussian process (GP) models \cite{ludkovski2018gaussian,huynh2021multi,huynh2021joint} provide a non-parametric spatio-temporal paradigm for longevity analysis within APC modeling. This approach runs parallel to the existing APC models and the newer Deep Learning-driven approaches \cite{nigri2019deep,perla2021time,richman2021neural}. The underlying prediction belongs to the class of \emph{spatial smoothers} and is similar to smoothing splines \cite{hastie1990generalized}. Among the main strengths of GPs are their flexibility, uncertainty quantification, and capabilities for multi-population analysis. Moreover, through their \emph{covariance kernel}, GPs offer a direct view into the inter-dependence of Age-Year-specific mortality rates, which enables the modeler to focus on capturing the respective covariance structure. The covariance kernel of a Gaussian process determines the properties of its distribution, including its posterior mean function, smoothness, and more. This offers valuable insight into the underlying dynamics of the process of interest, which is not possible with black-box methods like neural networks.

Matching the APC decomposition of the two-dimensional Age-Year mortality service into three univariate directions, one may consider kernels that reflect the Age structure of mortality, its evolution in time, and its cohort effects. In the existing GP mortality literature, this is straightforwardly translated into a \textit{separable} GP kernel---a product of a univariate kernel in Age, a univariate kernel in Year, and if desired, a univariate kernel in Birth Cohort. While offering a satisfactory performance, this choice is  quite restrictive and handicaps the ability of GPs to discover data-driven dependence.  With this motivation in mind, we explore GP kernel composition and discovery for mortality models. Our first goal in this article is thus to unleash an automated process for finding the covariance structures most appropriate for mortality analysis. One motivation is that different (sub-)populations have \emph{different} APC structures, and hence a one-size-fits-all approach is inadequate. To this end, we design a genetic programming algorithm that iteratively searches in the kernel space to find the most fit kernels. Our approach expands on the previous proposals for kernel discovery via Genetic Algorithms (GA), tailoring it to longevity analysis and introducing specialized mutation operations.
Along the way, we examine the GA's ability to recover a known APC structure through considering synthetic mortality surfaces, including its capability to identify cohort effects, and additive vs multiplicative effects.

Our second motivation is to link ideas in mortality modelling literature to the structures of different GP kernel families. We investigate a variety of kernels, vastly expanding upon the limited number of kernels (such as Squared Exponential and Mat\'ern) that have been considered for mortality so far. By introducing and testing new GP kernel families we remove the limitation of directly postulating the kernel family to be used, which leads to hidden restrictions and assumptions on the data.  Furthermore, additional kernels specifically represent richer structures including random walk, periodicity, and more general ARIMA processes. We observe that existing APC covariance structures, including the well known Lee-Carter \cite{lee2000lee} and Cairns-Blake-Dowd \cite{cairns2011mortality} families, can be exactly matched through additive GP kernels.
By testing various kernels, one can find better fits for a mortality surface and answer related questions, for example involving the strength or structure of a cohort effect.

Our core approach to above is compositional kernel search. Kernel composition utilizes the fact that kernels are closed under addition and multiplication. In turn, compositional kernels offer a rich and descriptive structure of underlying mortality dynamics.   The workhorse of our analysis is a Genetic Algorithm (GA) that uses the concept of generations to gradually discover better-and-better kernels through a mutation-selection mechanism.  Given a mortality dataset, the Genetic Algorithm described below generates vast quantities (in the thousands) of potential kernels. These kernels are sequentially fitted to the dataset and ranked according to a statistical fitness function. The GA then probabilistically promotes exploration of the most fit kernels and discards less fit ones. This procedure allows to automate the exploration of the best-performing GP models for mortality modeling.  Early proposals for compositional kernels with GPs involved forward search minimizing the Bayesian Information Criterion (BIC) to construct tree-based representations of kernels \cite{duvenaud2013structure,duvenaud2014automatic}. \citet{jin2020compositional} and \citet{roman2021evolving} build upon the idea using a GA; the former analyzes performance on several multi-dimensional synthetic test functions, and the latter on univariate time series. We extend and tailor these strategies for mortality modelling.

The compositional GP search framework fits naturally with mortality modelling.  A kernel can be viewed as a covariance function of a stochastic process, and in an APC application, the search is done through addition and multiplication of Age, Period, and Cohort effects. Such compositions and modifications are already performed in the aforementioned Lee-Carter and Cairns-Blake-Dowd (CBD) models (see \cite{cairns2011mortality} for thorough discussion).  Armed with the outputs of the GA,  we  address the following fundamental questions about mortality surfaces, which are of intrinsic interest:
\begin{itemize}
    \item The presence, or lack thereof, of a \emph{Cohort effect}. Our method offers a rigorous Bayesian nonparametric evidence on whether including Birth Cohort effect is beneficial. Since cohort effects are known to be population-specific, this is an important \emph{model selection} question.
    \item The relative smoothness between the Age- and Year- covariance structures. Classic APC models assume a random-walk structure in calendar time, and (implicitly) a smooth (infinitely-differentiable) structure in Age. In contrast, existing GP models have postulated a fixed smoothness (e.g.~twice differentiable) in both coordinates. Our method sheds light on whether those assumptions impact predictions and how much smoothness is most consistent with mortality data.

    \item Additive vs multiplicative structure in mortality covariance. There have been many proposals and comparative analyses of APC models that variously combine Age and Year terms. We provide an analogous analysis for GP models. In particular, our approach is able to quantify the \emph{complexity} of the best-fitting kernels, giving new insights about how many different terms are necessary.

\end{itemize}

The rest of the article is organized as follows. In Section \ref{sec:model} we review GP models for mortality surfaces, emphasizing the primordial task of kernel selection and illustrating its impact on model predictions. In Section \ref{sec:ga} we develop the Genetic Algorithm tool for compositional kernel search. Validity of the GA methodology is asserted in Section \ref{sec:synthetic} through a recovery of known kernels on synthetic mortality surfaces.  Section \ref{sec:hmd-results} analyzes the output of GA for the initial case study of JPN Females. Section \ref{sec:more-kernel-results} then provides a cross-sectional analysis across multiple national-level data sets to address the two questions of presence/absence of Cohort effects and the relative smoothness in Age and Calendar Year. Section \ref{sec:conclude} concludes.

\section{Gaussian Process Models for Mortality} \label{sec:model}

A \emph{Gaussian process} (GP) is a collection of random variables $\{f(\mathbf{x})\}_{\mathbf{x} \in \mathbb{\R}^d}$, such that for any $\ell \in \N$ and $\{\bx_1, \ldots, \bx_\ell\} \subseteq \R^d$, the vector $\left[f(\mathbf{x}_1),f(\mathbf{x}_2),...f(\mathbf{x}_\ell)\right]^\top$ has a multivariate normal distribution \cite{williams2006gaussian} (denoted $\text{MVN}$).  For mortality surfaces, $d=3$.  A GP is uniquely defined by its mean function $m: \R^d \rightarrow \R$ and covariance kernel $k : \R^d \times \R^d \rightarrow \R$ \cite{adler2010geometry}.  The kernel $k(\cdot, \cdot)$ must be a symmetric positive-definite function.  In this case, for $\mathbf{x}_i, \mathbf{x}_j \in \R^d$,
\begin{align}
    \mathbb{E}[f(\mathbf{x}_i)] &= m(\mathbf{x}_i),\\
    \text{cov}\left(f(\mathbf{x}_i),f(\mathbf{x}_j)\right) &= k(\mathbf{x}_i,\mathbf{x}_j),
\end{align}

\noindent and we write $f \sim \GP(m, k)$. The GP regression model assumes

\begin{equation}
    y := y(\bx) = f(\bx) + \eps(\bx), \label{eq:gpr}
\end{equation}
where $f$ is a Gaussian process with prior mean $m(\cdot)$ and covariance $k(\cdot, \cdot)$, $\eps$ is a noise term, and $y$ is a noisy observation.  By assuming $\eps(\cdot)$ is independent Gaussian white noise with variance $\sigma^2(\cdot)$, properties of multivariate normal random variables imply that $\{y(\bx)\}_{\bx \in \R^d}$ is a Gaussian process with mean and covariance functions
\begin{equation}\label{eq:y-mean-var}
\E[y(\bx)] = m(\bx), \qquad k_y(\bx_i, \bx_j') = k(\bx_i, \bx_j) + \sigma^2(\bx_i) \delta_{i = j},
\end{equation}
where $\delta$ is the Dirac delta. It is important to distinguish that  $\sigma^2(\bx_i) \delta_{i = j}$ is nonzero when the \emph{indices} $i$ and $j$ are equal: it is possible to have two observations at the same location $\bx$ but coming from different samples, thus not sharing noise.

\subsection{Gaussian Process Regression}\label{sec:gpr}

Given a data set $\cD = \{\bx_i, y_i\}_{i=1}^n$, the GP assumption and observation likelihood \eqref{eq:gpr} imply $[\mathbf{f}, \mathbf{y}]^\top \sim \MVN$, where $\mathbf{f} = [f(\bx_1), \ldots, f(\bx_n)]^\top$ and $\mathbf{y} = [y(\bx_1), \ldots, y(\bx_n)]^\top$, so that the posterior $\mathbf{f} | \mathbf{y} \sim \MVN$ as well.  More generally, for $\bx, \bx' \in \R^d$, $[f(\bx), f(\bx'), \mathbf{y}]^\top \sim \MVN$, so that the posterior finite dimensional distribution is fully known:

\begin{equation}
    [f_*(\bx), f_*(\bx')]^\top := \left([f(\bx), f(\bx')]^\top | \mathbf{y}\right) \sim \MVN \left([m_*(\bx), m_*(\bx')]^\top, \begin{bmatrix} k_*(\bx, \bx) & k_*(\bx, \bx') \\ k_*(\bx', \bx) & k_*(\bx', \bx') \end{bmatrix}\right), \label{eq:posterior-dist}
\end{equation}

where, for arbitrary $\bx, \bx' \in \R^d$, the posterior covariance kernel is defined as $k_*(\bx, \bx') := \cov(f(\bx), f(\bx') | \mathbf{y})$. The Kolmogorov extension theorem ensures that $\{f_*(\bx)\}_{\bx \in \R^d}$ defines a Gaussian process. Furthermore, the \emph{posterior mean} and \emph{variance} are explicitly given by
\begin{align}
    m_*(\bx) &= \mathbf{K}(\bx, \mathbf{X})\left[\mathbf{K}(\mathbf{X}, \mathbf{X}) + \Delta(\mathbf{X}, \mathbf{X})\right]^{-1}\mathbf{y},\\
    \mathbf{K}_*(\bx, \bx') &= \mathbf{K}([\bx, \bx']^\top, \mathbf{X}) \left[\mathbf{K}(\mathbf{X}, \mathbf{X}) + \bm{\Delta}(\mathbf{X}, \mathbf{X})\right]^{-1} \mathbf{K}(\mathbf{X}, [\bx, \bx']^\top),
\end{align}
where $\mathbf{X}$ denotes the $n \times d$ matrix with rows $\bx_i, i=1, \ldots, n$, and for $\mathbf{U}, \mathbf{V}$ being $\ell \times d$ and $m \times d$ respectively, $\mathbf{K}(\mathbf{U}, \mathbf{V})=\left[k(\mathbf{u}_i, \mathbf{v}_j)\right]_{1 \leq i \leq \ell, 1 \leq j \leq m}$ denotes the $\ell \times m$ matrix of pairwise covariances. $\bm{\Delta}(\mathbf{U}, \mathbf{V})$ has entries $\sigma^2(\mathbf{u}_i) \delta_{i = j}\delta_{\mathbf{u}_i = \mathbf{v}_j}$; in our case $\bm{\Delta}(\mathbf{X}, \mathbf{X})$ is a $n \times n$ diagonal matrix with entries $\sigma^2(\bx_i)$.  In the case of constant noise variance $\sigma^2 := \sigma^2(\bx_i)$, $\bm{\Delta}(\mathbf{X}, \mathbf{X}) = \sigma^2 I_n$, where $I_n$ is the $n \times n$ identity matrix.

\subsection{GP Kernels}

Gaussian process regression encodes the idea that similar inputs (according to the kernel) yield similar outputs. This can be seen through the posterior mean being a weighted average of observed data, since $m_*(\bx) = \mathbf{w}^\top \mathbf{y}$ holds for $\mathbf{w}^\top = \mathbf{K}(\bx, \mathbf{X})\left[\mathbf{K}(\mathbf{X}, \mathbf{X}) + \Delta(\mathbf{X},\mathbf{X})\right]^{-1}$. Various types of kernels exist to encode similarity according to domain knowledge. Akin to covariance matrices, the only requirement for a function $k(\cdot, \cdot)$ to be a covariance kernel is that it is symmetric and \emph{positive-definite}, i.e.~for all $n = 1, 2, \ldots,$ and $x_1, \ldots, x_n \in \R^d$, we must have the \emph{Gram matrix} $\mathbf{K}(\bX, \bX)$ be positive semi-definite.

A \emph{stationary} kernel $k(\mathbf{x}_i, \mathbf{x}_j)$ is one that can be written as a function of $\mathbf{x}_i - \mathbf{x}_j$, that is, $k(\mathbf{x}_i, \mathbf{x}_j) = k_S(\mathbf{x}_i - \mathbf{x}_j) $ and is thus invariant to translations in the input space. A kernel is further called \emph{isotropic} if it is only a function of  $\mathbf{r} = \Vert\mathbf{x}_i - \mathbf{x}_j\Vert$, where $\Vert \cdot \Vert$ is the $\ell^2$ Euclidean distance, so that we can write $k_I(\mathbf{r}) = k(\mathbf{x}_i, \mathbf{x}_j)$.  \citet{yaglom1957some} uses Bochner's theorem to derive a similar Fourier transform specific to isotropic kernels, providing a way to derive kernels from spectral densities. Stationary kernels are usually assumed to be normalized, since $k_S(\bx-\bx')/k_S(\mathbf{0})=1$ whenever $\bx = \bx'$; this allows for stationary kernels to be nicely interpreted as correlation functions.  Lastly, a \emph{separable} kernel over $\R^d$ is one that can be written as a product: $k(\bx, \bx') = \prod_{j=1}^d k_j(x^{(j)}, x^{'(j)})$, where $x^{(j)}$ is the $j$th coordinate of $\bx$.  Thus, the global kernel is separated as a product over its dimensions, each having its own kernel.

The algebraic properties of positive-definite functions make it straightforward to compose new kernels from existing ones \cite{scholkopf2002learning,shawe2004kernel, berlinet2011reproducing,genton2001classes}. The main tool is that kernels are preserved under addition and multiplication, i.e.~can be combined by sums and products. Hence, if $k_1$ and $k_2$ are two kernels and $c_1, c_2$ are two positive real numbers, then so is $k(\bx, \bx') = c_1 k_1(\bx, \bx') + c_2 k_2(\bx, \bx')$. This is consistent with the properties of Gaussian processes: if $f_1\sim \GP(0, k_1)$ and $f_2 \sim \GP(0, k_2)$ are two independent GPs, then for $c_1, c_2 >0$ we have $c_1 f_1+c_2 f_2 \sim \GP(0, c_1 k_1 + c_2 k_2)$. This offers a connection to the framework of generalized additive models.  Although there is no analogous result for a product of kernels (a product of GPs is no longer a GP, since the multivariate Gaussian distribution is not preserved), a common interpretation of multiplying kernels occurs when one kernel is stationary and monotonically decaying as $|\bx-\bx'| \rightarrow \infty$.  Indeed, if $k_1$ is such a kernel, then the product $k_1(\bx,\bx')k_2(\bx,\bx')$ offers $k_2$'s effect with a decay according to $k_1(\bx,\bx')$ as $|\bx-\bx'|$ increases. This synergizes with separability, for example $k(\bx, \bx') = k_1(|x^{(1)} - x^{'(1)}|)k_2(x^{(2)}, x^{'(2)})$ offers some similarity across the second coordinate that could decay as $|x^{(1)} - x^{'(1)}|$ increases.  Additive and multiplicative properties are often used in conjunction with the constant kernel $k(\bx, \bx') = c, c >0$, resulting in a scaling effect (multiplication) or dampening effect (addition).

\begin{rem}
    Several other kernel design strategies exist, for example if $g : \R^d \rightarrow \R$, then $k(\bx, \bx') = g(\bx)g(\bx')$ defines a kernel; this property along with multiplication is one way to account for heteroskedastic noise, since $k(\bx_i, \bx'_j) = \delta_{i=j}$ defines a kernel. See e.g.~\citet{genton2001classes} or \citet{noack2021advanced} for additional properties and explanation.
\end{rem}

\subsection{Kernel Families}\label{sec:kernel_details}

Table \ref{tab:kernels} lists the kernel families we consider. For simplicity, we assume one dimensional base kernels with $x \in \R$ where the full structure for $\bx \in \R^d$ is expressed as a separable kernel.  Among the 9 kernel families, we have kernels that give smooth ($C^2$ and higher) fits, kernels with rough (non-differentiable) sample paths, and several non-stationary kernels. Here, smoothness refers to the property of the sample paths $\bx \mapsto f(\bx)$ being, say, $k$-times differentiable, i.e.~$f(\cdot) \in C^k$.  The posterior mean $\bx \mapsto m_*(\bx)$ inherits similar (but generally less strict) differentiability properties; see \cite{kanagawa2018gaussian} for details.
The $\mathcal{K}_r$ column indicates whether a kernel is included in the \emph{restricted set of kernels} used in later sections.  This comprises a more compact collection of the most commonly used kernels in the literature. One reason for ${\cal K}_r$ is to minimize overlap in terms of kernel properties; as we report below, some kernel families in ${\cal K}_f$ apparently yield very similar fits, and act as ``substitutes'' for each other.

\begin{table}[]
    \centering
\begin{tabular}{rc|lp{1in}c} \toprule
Kernel Name & Abbv. & Formula $k(x, x'; \theta)$  & Properties  & $\mathcal{K}_r$\\ \midrule
 Mat\'ern-1/2 & $\M12$   & $\exp\Big(-\frac{|x-x'|}{\ell_{\mathrm{len}}}\Big), \quad \ell_{\mathrm{len}} > 0$ & $C^0$ & \checkmark \\
 Mat\'ern-3/2 & $\M32$    & $\left(1 + \frac{\sqrt{3}}{\ell_{\mathrm{len}}}|x - x'|\right) \exp\Big(\!-\frac{\sqrt{3}}{\ell_{\mathrm{len}}}|x - x'|\Big), \quad \ell_{\mathrm{len}} > 0$ &  $C^1$ &  \\
 Mat\'ern-5/2 & $\M52$    & $\left(1 + \frac{\sqrt{5}}{\ell_{\mathrm{len}}}|x - x'| + \frac{5}{3\ell^2_{\mathrm{len}}}|x - x'|^2\right) \exp\Big(\!-\frac{\sqrt{5}}{\ell_{\mathrm{len}}}|x - x'|\Big)$ &  $C^2$ & \checkmark \\
 Cauchy & $\Chy$ & $\frac{1}{1+ |x-x'|^2/\ell_{\mathrm{len}}^2}, \quad \ell_{\mathrm{len}} > 0$ & $C^\infty$ & \\
Radial Basis & $\RBF$  & $\exp{\Big(\!- \frac{(x - x' )^2  }{2 \ell^2_{\mathrm{len}}}\Big)}, \quad \ell_{\mathrm{len}} > 0$  & $C^\infty$ & \checkmark\\
 AR2 & AR2 & $\exp(-\alpha |x-x'|)\left\{\cos(\omega |x-x'|) + \frac{\alpha}{\omega} \sin(\omega |x-x'|)\right\}$ & Periodic, $C^1$ &  \\
 Linear & Lin & $\sigma_0^2 + x \cdot x', \quad \sigma_0 > 0$ & Non-stationary & *\\
 Minimum & $\Min$ & $t_0^2 + x \wedge x', \quad t_0 > 0$ & Non-stat, $C^0$ & \checkmark\\
Mehler & $\Meh$ & $ \exp\left(-\frac{\rho^2(x^2 + x'^2) - 2 \rho xx'}{2(1-\rho^2)}\right), \quad -1 \leq \rho \leq 1$ & Non-stationary  & \\ \bottomrule
 \end{tabular}
    \caption{List of kernel families used in compositional search. $C^p$ indicates that the GP sample paths $x \mapsto f(x)$ have $p$ continuous derivatives; $C^0$ is continuous but not differentiable.  Column $\mathcal{K}_r$ denotes whether the kernel family is in the restricted search set. The linear kernel is used for its year component only. 
    }
    \label{tab:kernels}
\end{table}

All of the kernels listed have hyperparameters which help to understand their relationship with the data.  The quantity $\ell_{\text{len}}$ appearing in many stationary kernels is referred to as the \emph{characteristic length-scale}, which is a distance-scaling factor.  With the RBF kernel for example, this loosely describes how far $x'$ needs to move from $x$ in the input space for the function values to become uncorrelated \cite{williams2006gaussian}. Thus, the lengthscale of a calibrated GP can be  interpreted as the strength of the correlation decay in the training dataset.  Out of the stationary kernels, a popular class is the Mat\'ern class.  In continuous input space, the value $\nu$ in the Mat\'ern-$\nu$ corresponds to smoothness: a GP with a Mat\'ern-$\nu$ kernel is $\lceil \nu \rceil -1$ times differentiable in the mean-square sense \cite{williams2006gaussian}.  The Radial Basis Function (RBF) kernel is the limiting case as $\nu \rightarrow \infty$, resulting in an infinitely differentiable process.  The $\nu=1/2$ case recovers the well known Ornstein-Uhlenbeck process, which is mean reverting and non-differentiable.  Also non-differentiable but on the nonstationary side, the minimum kernel corresponds to a Brownian motion process when $x \in \R_+$, where $t_0^2 = \var(f(0))$ is the initial variance. For discrete $x$, $\Min$ kernel yields the random walk process, and $\M12$ yields the AR(1) process.

Less commonly studied are the (continuous-time) AR2 (see \citet{parzen1961approach}), Cauchy, and Mehler kernels.  The continuous-time AR2 kernel acts identically to a discrete-time autoregressive (AR) process of order 2 with complex characteristic polynomial roots when $x$ is restricted to an integer.
The heavy tailed Cauchy probability density function motivates the Cauchy kernel, with the goal of modelling long-range dependence and is a special case of the rational quadratic kernel (see Appendix \ref{sec:kernel_appendix}).  Lastly, the Mehler kernel has a form similar to that of a joint-normal density, and acts as a $\RBF$ kernel with a non-stationary modification (this can be seen from a ``complete the squares" argument).  Although Mehler is non-stationary, it remarkably yields a stationary \emph{correlation function} $\text{corr}(f(\bx), f(\bx')) = k(\bx, \bx') / \sqrt{k(\bx, \bx) k(\bx', \bx')}$. See Appendix \ref{sec:kernel_appendix} for a more thorough discussion of the aforementioned kernels and their properties.

\subsection{Connections to Mortality Modelling}

For mortality modeling, our core input space is composed of Age and Year coordinates: $x_{a}, x_{y} \in \mathbb{R}_+^2$.
As the GP can model non-linear relationships, we include birth cohort $x_c := x_a-x_y$ as a third coordinate of $\bx$, so that $\bx = (x_{a}, x_{y}, x_{c})$.   For a given $\bx$, denote $D_{\bx}$ and $E_{\bx}$ as the respective observed deaths and exposures (i.e.~individuals alive at the beginning of the period) over the corresponding $(x_{a}, x_{y})$ pair.  Denote $y(\bx) = \log(D_{\bx} / E_{\bx})$ as the \emph{log mortality rate}.  The full data set is denoted $\mathcal{D} = \{\bx_i, y_i, D_i, E_i\}_{i=1}^n$.  
Each mortality observation $y_i := y_i(\bx_i)$ is the regression quantity modelled in Equation \eqref{eq:gpr}, so that $\E[y(\bx) | f(\bx)] = f(\bx)$. The interpretation is that the true mortality rate $f(\bx)$ is observed in accordance with mean-zero (i.e.~unbiased) uncorrelated noise yielding the measured mortality experience $y$.

Relating to the log-normal distribution, algebra shows that
$$\E[D_{\bx} | f(\bx)] = E_{\bx} \exp\left(f(\bx) + \frac{\sigma^2(\bx)}{2}\right),  \text{ and } \quad \E[D_{\bx} | f(\bx), \eps(\bx)] = E_{\bx}\exp\left(f(\bx) + \eps(\bx)\right),$$
akin to an overdispersed Poisson model in existing mortality modelling literature (see e.g.~\cite{azman2022glm}).  Note that the seminal work of \citet{brouhns2002poisson} for modelling log-mortality rates suggests that homoskedastic noise is unrealistic, since the absolute number of deaths at older ages is much smaller compared to younger ages.  As a result, we work with heteroskedastic noise $$\var(\epsilon | D_{\bx}) = \sigma^2(\bx) := \frac{\sigma^2}{D_{\bx}}, \quad \text{where } \sigma^2 \in \R^+.$$
Thus, we make observation variance inversely proportional to observed death counts, with the constant $\sigma^2$ to be learned as part of the fitting procedure. From a modelling perspective this works since $D_{\bx}$ is known whenever $E_{\bx}$ and $y(\bx)$ are.  Indeed, Equation \eqref{eq:posterior-dist} shows no requirement to know $D_{\bx}$ for out-of-sample forecasting.  In the case where full distributional forecasts are desired, one could instead model a noise surface $\sigma^2(\cdot)$ simultaneously with $f(\cdot)$; see for example \cite{cole2022large}.

A discussion of GP covariances connects naturally to APC models.  A stationary covariance means that the dependence between different age groups or calendar years is only a function of the respective ages/year distances, and is not subject to additional structural shifts.
Additive and separable covariance structures play an important role in the existing mortality modelling literature, specifically in the APC framework.  For example, Lee-Carter and Cairns-Blake-Dowd (CBD) models began with Age-Period models, which subsequently evolved to add cohort or additional Period effects. Rather than postulating the precise APC terms, the latest
\citet{dowd2020cbdx} CBDX framework adds period effects as needed. Similarly, \citet{hunt2014general} develop a general recipe for constructing mortality models, where core demographic features are represented with a particular parametric form, and combined into a global structure.  This can be done with GP's where kernels encode expert judgment to such demographic features.  This type of encoding into covariances is already being done in the literature, possibly unknowingly.  Take for example the basic CBD model whose stochastic part has the form $f(\bx) = \kappa_{x_{yr}}^{(1)} + (x_{ag}-\overline{\bx_{ag}}) \kappa_{x_{yr}}^{(2)}$, where $\overline{\bx_{ag}}$ is the average age in $\mathcal{D}$.  Under the common assumption that $(\kappa_{x_{yr}}^{(1)}, \kappa_{x_{yr}}^{(2)})$ is a multivariate random walk with drift, a routine calculation shows that $$\E[f(\bx)] = \kappa_0^{(1)}+\mu^{(1)}x_{yr} + (x_{ag}-\overline{\bx_{ag}})(\kappa_0^{(2)}+\mu^{(2)}x_{yr})$$ and $ k_{CBD}(\bx, \bx') = \cov(f(\bx), f(\bx')) =$
\begin{equation}
     = \left[ \sigma_1^2 + \rho \sigma_1 \sigma_2(x_{ag} + x'_{ag} - 2\overline{\bx_{ag}}) + (x_{ag} -\overline{\bx_{ag}})(x'_{ag} - \overline{\bx_{ag}}) \sigma_2^2 \right]\left(x_{yr} \wedge x'_{yr}\right).
\end{equation}

We can re-interpret the above as the kernel decomposition $k_{CBD}(x, x') = k_1(x_{ag}, x'_{ag})k_2(x_{yr}, x'_{yr})$, where $k_2(x_{ag}, x'_{ag})$ is a kernel depending only on Age (that could be decomposed into additive components), and $k_2$ is the minimum kernel depending only on Year.  Furthermore, if the multivariate random walk is assumed to be Gaussian, then $f(\bx)$ actually forms a (discrete-time) Gaussian process, and hence the methods detailed in Section \ref{sec:gpr} apply verbatim.

Period and cohort effects are commonly modelled using time series models \cite{villegas2015stmomo}. In particular, Gaussian ARIMA models are popular in existing APC literature; see for instance \cite{cairns2011mortality} who single out the usefulness of AR(1), ARIMA(1,1,0) and ARIMA(0,2,1) for Cohort effect, and ARIMA(1,1,0) or ARIMA(2,1,0) for Period effect. This provides another link to (discrete) GP covariance analogues:  a Mat\'ern-1/2 covariance corresponds to an AR(1) process, a (continuous-time) AR2 covariance corresponds to an AR(2) process with complex unit roots, and a Minimum covariance to a discrete-time random walk.

\section{Genetic Programming for GPs} \label{sec:ga}

Starting with the building blocks of the kernels in Table \ref{tab:kernels}, infinitely many compositional kernels can be constructed through addition and multiplication.
The idea of a genetic algorithm is to adaptively explore the space of kernels via an evolutionary procedure. At each step of the GA, kernels that have a higher ``fitness score" are more likely to evolve and be propagated, while lower-fitness kernels get discarded. The evolution is achieved through several potential operations, that are selected randomly in each instance. In the first sub-step of the GA, \emph{ancestors} of next-generation kernels are identified. This is done via ``tournaments" that aim to randomly pick generation-$g$ kernels, while preferring those with higher fitness. A given kernel can be selected in multiple tournaments, i.e.~generate more than one child.  In the second sub-step, each ancestor undergoes crossover (mixing kernel components with another ancestor) or mutation (modifying a component of the sole ancestor) to generate a generation-($g+1$) kernel.

\subsection{BIC and Bayes Factors for GPs}

To evaluate the appropriateness of a kernel within a given set  $k \in \mathcal{K}$, an attractive criterion is the posterior likelihood of the kernel given the data $p(k | y) = p(y | k) p(y) / p(k)$, where, under a uniform prior assumption $p(k) = 1/|\mathcal{K}|$, we see that $p(k | y) \propto p(y | k)$. However, the integral over hyperparameters $p(y | k) = \int_\theta p(y, \theta| k)d\theta$ is generally intractable, so we use the \emph{Bayesian Information Criterion} ($\BIC$) as an approximation, where $\BIC(k) \approx \log p(y | k)$ is defined as:

\begin{equation}\label{eq:bic}
    \BIC(k) = -l_{k}(\hat{\theta}; \mathbf{y}) + \frac{|\hat{\theta}| \log(n)}{2},
\end{equation}

\noindent where $l_k(\theta| \mathbf{y}) = \log p(y | k, \theta)$ is the \emph{log marginal likelihood of $y$} evaluated at $\theta$ under a given kernel $k$, $\hat{\theta}$ is the maximizer (maximum marginal likelihood estimate) of $l_k(\theta| \mathbf{y})$, and $|\hat{\theta}|$ is the total number of estimated hyperparameters in $\hat{\theta} = \{\hat{\beta}_0, \hat{\beta}_{ag}, \hat{\theta}_k, \hat{\sigma}^2\}$, where $\theta_k$ is a vector of all kernel specific hyperparameters. Note that $p(y|k, \theta)$ is a multivariate density, with mean and covariances governed by Equation \eqref{eq:y-mean-var}.  The BIC metric has seen use in similar applications of GP compositional kernel search, see e.g.~\cite{duvenaud2013structure,duvenaud2014automatic}, and is commonly used in mortality modelling \cite{cairns2009quantitative}. We employ BIC (lower BIC being better)  for our GA fitness metric  below.

One can further assess the relative likelihood of $k_1, k_2 \in \mathcal{K}$ by again assuming a uniform prior over $\mathcal{K}$, and computing the \emph{Bayes Factor} (BF)
\begin{equation} \label{eq:bayes_factor_bf}
\text{BF}(k_1, k_2) = \frac{p(k_1 | \mathbf{y})}{p(k_2 | \mathbf{y})} \approx \exp\Big(\BIC(k_2) - \BIC(k_1)\Big).
\end{equation}

\citet{gelman1995bayesian} states that Bayes factors work well in the case of a discrete model selection.  The seminal work of \citet{jeffreys1961theory} gives a table of \emph{evidence categories} to determine a conclusion from Bayes factors; see Table \ref{tab:bf_evidence_categories} in Appendix \ref{sec:bf_appendix} which is still frequently used today \cite{lee2014bayesian, dittrich2019network}.

Note that in accordance with the penalty term $|\hat{\theta}| \log(n)/2$ in Equation \eqref{eq:bic}, the difference in penalties in  $\BIC(k_2) - \BIC(k_1)$ is simply the number of additional kernel hyperparameters, scaled by $\log(n)/2$, since $\beta_0, \beta_{ag}, \sigma^2$ are always estimated regardless of kernel choice. Thus in the application of GP kernel selection, the Bayes factor properly penalizes kernel complexity.  In Table \ref{tab:kernels} most kernels have one $\theta_k$ (the lengthscale), but some, like $\AR2$, have two hyperparameters and so incur a relative penalty.

\subsection{GA Kernel Representation}

In order to operate in the space of kernels, we shall represent kernels via
 a tree-like structure, cf.~Figure \ref{fig:GA_operations} .  Internal nodes correspond to \emph{operators} (\texttt{add} or \texttt{mul}) that combine 2 different kernels together, while leafs are the univariate kernels used as building blocks. We indicate the coordinate operated on by the kernel through the respective subscripts $a,y,c$, such as $\M52_{a}$. In turn, such trees are transcribed into bracketed expressions, such as $\kappa = \texttt{add(Exp\_c, mul(RBF\_a, add(Mat\_y, RBF\_c)))}$
corresponding to the Age-Period-Cohort kernel $(k_{M52}(x_{yr})+k_{RBF}(x_c) )\cdot k_{RBF}(x_{ag}) + k_{Exp}(x_c)$. The \emph{length} of $\kappa$, denoted $|\kappa|$, is its number of nodes. The above kernel tree has length $|\kappa| = 7$, namely 4 base kernels combined with 3 operators. Observe that the tree structure is not unique, i.e.~some complex kernels can be permuted and expressed through different trees. In what follows we will ignore this non-uniqueness.

A certain expertise is needed to convert from an above representation to the dependence structure it implies. One way to visualize is to take advantage of the stationarity and plot the heatmap of the matrix $k_S(\bx-\bx')$ as a function of its Age, Year coordinates.

\subsection{GA Operations and Hyperparameters}

The overall GA is summarized by the set of possible mutations and a collection of hyperparameters.  These are important for many reasons: (i) sufficient exploration so that the algorithm does not get trapped in a particular kernel configuration; (ii) efficiency in terms of number of generation and generation size needed to find the best kernels; (iii) bloat control, i.e.~ensuring that returned kernels are not overly complex and retain interpretability. Interpretable kernels would tend to have low length (below 10) and avoid repetitive patterns.  Bloat control, i.e.~avoiding the appearance of overly complex/long kernels is a concern with GAs. \citet{luke2006comparison} and \citet{o2009riccardo} suggest several ways to combat it.

Our specific high-level GA hyperparameters are listed in Table \ref{tab:ga_params_high_level}. We largely follow guidelines from \citet{sipper2018investigating} which offers a thorough investigation of the parameter space of GA algorithms.  Ancestors are chosen via a {tournament} setup, where $T=7$ individuals are independently and uniformly sampled from the previous generation and a single {tournament winner} is the {fittest} (lowest BIC) individual.  A smaller $T$ provides diversity in future generations, whereas a larger $T$ reduces chances of leaving behind fit individuals. To combat bloat, we follow the \emph{double tournament} procedure described in \cite{luke2006comparison}: all instances of a single tournament are replaced by two tournaments run one after the other, with the lower length ancestor chosen with probability $D/2$ ($D \in [1, 2]$; larger values prefer parsimony over fitness).  We use the less restrictive $D=1.2$ instead of the suggested $D=1.4$, partially since BIC has a built-in penalization for unwieldy models thereby mitigating bloat.  As a further proponent of parsimony, we use hoist mutation as suggested in \citet{o2009riccardo} with $p=0.1$. Using the above parameterization, we rarely observe kernels of length over 15 in our experiments.

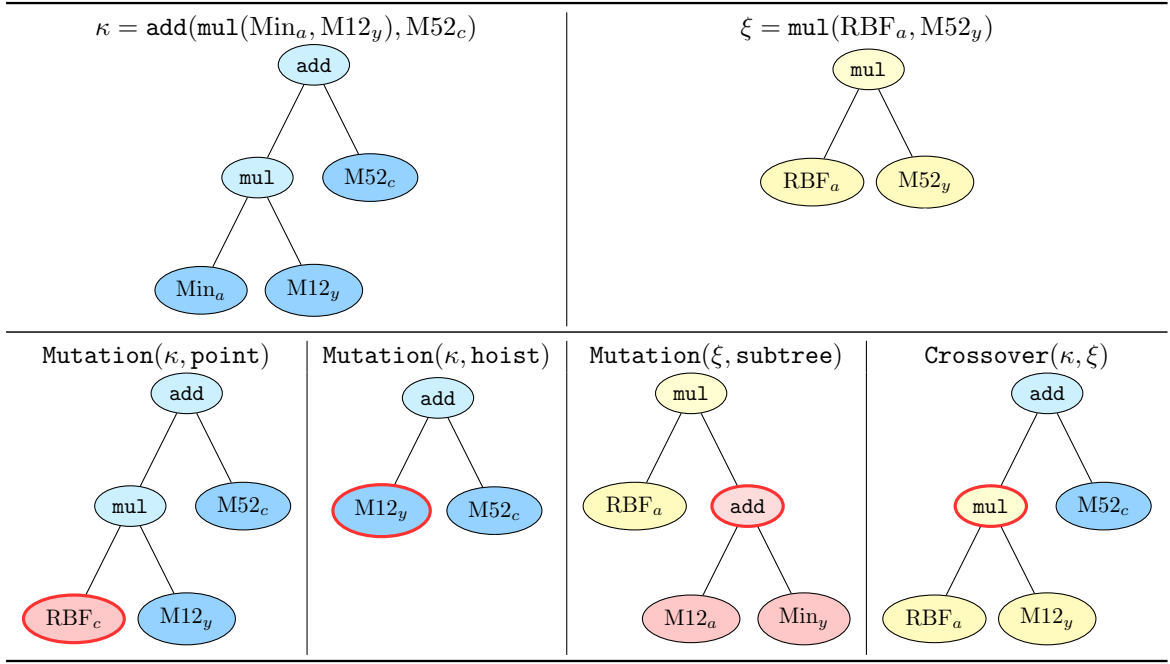
\begin{figure} {\small
    \centering
\begin{tabular}{c|c|c|c} \toprule
     \multicolumn{2}{c|}{$\kappa=\texttt{add}(\texttt{mul}(\Min_a, \M12_y),\M52_c)$} & \multicolumn{2}{c}{$\xi=\texttt{mul}(\RBF_a,\M52_y)$} \\
     \multicolumn{2}{c|}{
\begin{tikzpicture}
[emph/.style={edge from parent/.style={white,very thick,draw}},
    norm/.style={edge from parent/.style={black,thin,draw}}]
\node [ellipse,draw,fill=blue1] (z){$\texttt{add}$}
  child {node [ellipse=red1,draw,fill=blue1] (a) {$\texttt{mul}$}
    child {node [ellipse,draw,fill=blue2] (b) {$\Min_a$}}
    child {node [ellipse,draw,fill=blue2] (c) {$\M12_y$}}
  }
  child {node [ellipse,draw,fill=blue2] (d) {$\M52_c$}
};
\end{tikzpicture}
     }
     &
     \multicolumn{2}{c}{
\begin{tikzpicture}
[emph/.style={edge from parent/.style={white,very thick,draw}},
    norm/.style={edge from parent/.style={black,thin,draw}}]
\node [ellipse,draw,fill=yellow1] (z){$\texttt{mul}$}{
    child {node [ellipse,draw,fill=yellow2] (b) {$\RBF_a$}}
    child {node [ellipse,draw,fill=yellow2] (c) {$\M52_y$}
      child[emph] {node [ellipse, draw=white, color = white] (d) {$\textcolor{white}{AAA}$}}
    }
};
\end{tikzpicture}
     }    \\ \midrule
    $\texttt{Mutation}(\kappa, \texttt{point})$ & $\texttt{Mutation}(\kappa, \texttt{hoist})$ & $\texttt{Mutation}(\xi, \texttt{subtree})$ & $\texttt{Crossover}(\kappa, \xi)$\\
    \begin{tikzpicture}
[emph/.style={edge from parent/.style={white,very thick,draw}},
    norm/.style={edge from parent/.style={black,thin,draw}}]
\node [ellipse,draw,fill=blue1] (z){$\texttt{add}$}
  child {node [ellipse,draw,fill=blue1] (a) {$\texttt{mul}$}
    child {node [ellipse,draw=red3,fill=red2,text=red4,very thick] (b) {$\RBF_c$}}
    child {node [ellipse,draw,fill=blue2] (c) {$\M12_y$}}
  }
  child {node [ellipse,draw,fill=blue2] (d) {$\M52_c$}
};
\end{tikzpicture} &

\begin{tikzpicture}
[emph/.style={edge from parent/.style={white,very thick,draw}},
    norm/.style={edge from parent/.style={black,thin,draw}}]
\node [ellipse,draw,fill=blue1] (z){$\texttt{add}$}{
    child {node [ellipse,draw=red3,fill=blue2,very thick] (b) {$\M12_y$}}
    child {node [ellipse,draw,fill=blue2] (c) {$\M52_c$}
      child[emph] {node [ellipse, draw=white, color = white] (d) {$\textcolor{white}{AAA}$}}
    }
};
\end{tikzpicture} &

\begin{tikzpicture}
[emph/.style={edge from parent/.style={white,very thick,draw}},
    norm/.style={edge from parent/.style={black,thin,draw}}]
\node [ellipse,draw,fill=yellow1] (z){$\texttt{mul}$}{
  child {node [ellipse,draw,fill=yellow2] (a) {$\RBF_a$}}
  child {node [ellipse,draw=red3,fill=red1,very thick] (b) {$\texttt{add}$}
    child {node [ellipse,draw,fill=red2] (c) {$\M12_a$}}
    child {node [ellipse,draw,fill=red2] (d) {$\Min_y$}}
  }
};
\end{tikzpicture} &

\begin{tikzpicture}
[emph/.style={edge from parent/.style={white,very thick,draw}},
    norm/.style={edge from parent/.style={black,thin,draw}}]
\node [ellipse,draw,fill=blue1] (z){$\texttt{add}$}
  child {node [ellipse,draw=red3,fill=yellow1,very thick] (a) {$\texttt{mul}$}
    child {node [ellipse,draw,fill=yellow2] (b) {$\RBF_a$}}
    child {node [ellipse,draw,fill=yellow2] (c) {$\M12_y$}}
  }
  child {node [ellipse,draw,fill=blue2] (d) {$\M52_c$}
};
\end{tikzpicture}\\ \bottomrule

\end{tabular}
} 

    \caption{Representative compositional kernels and GA operations.  Bolded red ellipses indicate the node of $\kappa$ (or $\xi$) that was chosen for mutation or crossover.}
    \label{fig:GA_operations}
\end{figure}

Table \ref{tab:ga_params_high_level_2} fully details the crossover and mutation operations and their hyperparameters, with Figure \ref{fig:GA_operations} providing a visual illustration.  Most of these are standard in the literature. Our domain knowledge suggests an additional point mutation operator which we call \emph{respectful point mutation}. This operator maintains the coordinate of the kernel being mutated, so that a kernel operating on age is replaced by one in age, and so forth. This respects a discovered APC structure, and fine tunes the chosen coordinate.

\tikzstyle{every node}=[ellipse, scale=0.85, align=center]

The \emph{arity} (number of arguments) of a function needs to be preserved in crossover and mutation operations.  In our setup, the only non-trivial arity functions are \texttt{add} and \texttt{mul} (both with arity 2), so if these nodes are chosen for a mutation, the point mutation (and respectful point mutation) automatically replaces them with the other operation: \texttt{add} with \texttt{mul} and vice-versa.

In some GA applications, authors propose to have several hundred (or even thousand) generations. Because times to fit a GP model is non-trivial, running so many generations is computationally prohibitive. Below we use a fixed number of $G=20$ generations. As shown in Figure \ref{fig:ga_convergence}, kernel exploration seems to stabilize after a dozen or so generations, so there appears to be  limited gain in increasing $G$.  In contrast with a typical GA application, we are more interested in interpreting all prototypical well-performing kernels, rather than in optimizing an objective function to an absolute minimum; that is to say, our analysis is sufficient as long as a representative ballpark of optimal models has been discovered.

\begin{table}
    \centering
    \begin{tabular}{rr|p{3.5in}} \toprule
     Parameter & Value & Notes \\ \midrule
        Population Size & $n_g=200$ & Number of individuals per generation \\
        Generations & $G = 20$ & Number of generations \\
        Tournament Size & $T = 7$ & Run a double tournament and select smaller winner with probability $D/2$\\
        & $D = 1.2$ & Smaller is winner with probability $0.60$  \\ \bottomrule
    \end{tabular}
    \caption{High level Genetic algorithm hyperparameters and description.  Note that $G \cdot n_g = 4000$ is the total number of trained GP models in a single run of the GA.}
    \label{tab:ga_params_high_level}
\end{table}

\begin{table}
    \centering
    \begin{tabular}{l|p{1.5in}|p{3.5in}} \toprule
       Probability & GA Operation & Notes \\ \midrule
        $p_c = 0.45$ & Crossover & \\
        $p_s = 0.2$ & Subtree Mutation &\\
        $p_h = 0.1$ & Hoist Mutation &\\
        $p_p = 0.05$ & Point Mutation & Each node is mutated with another node of same arity with prob.~$q_p$\\
        $p_r = 0.15$ & Respectful Point Mutation & Each node is mutated with another node of same arity and same $(\text{age}, \text{year}, \text{cohort})$ with prob.~$q_r$\\
        $p_o = 0.05$ & Copy &\\ \midrule
        $q_p = 0.25$ & Point Replace  &\\
        $q_r = 0.35$ & Respectful Replace \\
        $q_a = 0.5$ & & Probability that \texttt{add}/\texttt{mul} is included when initializing trees \\ \bottomrule
    \end{tabular}

    \caption{Operator specific GA hyperparameters with description.  Note that $p_c + p_s + p_h + p_p + p_r + p_0 = 1$.}
    \label{tab:ga_params_high_level_2}
\end{table}

\IncMargin{1em}
\begin{algorithm}
\SetKwData{Left}{left}\SetKwData{This}{this}\SetKwData{Up}{up}
\SetKwFunction{Union}{Union}\SetKwFunction{FindCompress}{FindCompress}
\SetKwFunction{Union}{Union}\SetKwFunction{BIC}{BIC}
\SetKwFunction{Union}{Union}\SetKwFunction{SampleUniform}{SampleUniform}
\SetKwFunction{Union}{Union}\SetKwFunction{Sample}{Sample}
\SetKwFunction{Union}{Union}\SetKwFunction{InitializeKernel}{InitializeKernel}
\SetKwFunction{Union}{Union}\SetKwFunction{Operation}{Operation}
\SetKwFunction{Union}{Union}\SetKwFunction{Mutation}{Mutation}
\SetKwFunction{Union}{Union}\SetKwFunction{Type}{Type}
\SetKwFunction{Union}{Union}\SetKwFunction{Crossover}{Crossover}
\SetKwBlock{DetermineAncestor}{DetermineAncestor}{DetermineAncestor}
\SetKwInOut{Input}{input}\SetKwInOut{Output}{output}
\Input{Kernel search set $\cK$; genetic algorithm hyperparameters $\mathcal{G}_1 = \{n_g, G, T\}$ and $\mathcal{G}_2$ in Tables \ref{tab:ga_params_high_level} and \ref{tab:ga_params_high_level_2} respectively; Data set $\mathcal{D} = \{(\bx_1, y_1), \ldots, (\bx_n, y_n)\}$}
\Output{Kernel expressions and BIC scores $\{\{\kappa_i^g, b_i^g\}_{i=1}^{n_g}\}_{g=0}^{G-1}$}
\BlankLine
\For{$i= 1$ \KwTo $m_0$}{
    $\kappa_i^0 \leftarrow$ \InitializeKernel{}\;
    $b_i^0 \leftarrow$ \BIC{$\kappa_i^0$}
}
\For{$g= 1$ \KwTo $G-1$}{
    \For{$i= 1$ \KwTo $n_g$}{
        \DetermineAncestor{
        $\{\kappa^{g-1}_{(1)}, \ldots, \kappa^{g-1}_{(T)}\} \leftarrow $ \SampleUniform{$\{\kappa_1^{g-1}, \ldots, \kappa_{n_{g-1}}^{g-1}\}, T$} \;
        $j_i \leftarrow \text{argmin}\{b^{g-1}_{(1)}, \ldots, b^{g-1}_{(T)}\}$\;
        $A_i^{g-1} \leftarrow \kappa^{g-1}_{j_i}$    \;
        }
        \Operation $\leftarrow $ \Sample{\Crossover, (\Mutation, \Type)} \tcp*[f]{according to $\mathcal{G}_2$}\;
        \If{\Operation == \Crossover}{
        ${A'}^{g-1}_i \leftarrow \kappa^{g-1}_{j'_i}$ \tcp*[f]{2nd ancestor (according to \textbf{DetermineAncestor})}\;
        $\kappa_i^g \leftarrow $ \Crossover{$A_i^{g-1}, {A'}^{g-1}_i$}\;
        }
        \Else{
        $\kappa_i^g \leftarrow $ \Mutation{$A_i^{g-1}$; \Type}\;
        }
        $b_i^g \leftarrow$ \BIC{$\kappa_i^g$}
    }
}
\caption{Genetic Algorithm for compositional kernel selection}\label{algo:GA_algo}
\end{algorithm}\DecMargin{1em}

Denote by $\kappa_i^g$ the $i = 1, \ldots, n_g$th \emph{individual} in generation $g=1, \ldots, G-1$ as $\kappa_i^g$, with a corresponding BIC of $b_i^g$.
Algorithm \ref{algo:GA_algo} outlines the full genetic algorithm.   \texttt{BIC} computes the BIC as in Equation \eqref{eq:bic}, which implicitly fits a GP and optimizes hyperparameters.  \texttt{SampleUniform} determines ancestors according to a {tournament} of size $T$.  A crossover or mutation is determined according to the probabilities provided in $\mathcal{G}_2$ (according to Table \ref{tab:ga_params_high_level_2}), where \texttt{Type} denotes the type of mutation chosen. In the event of a crossover, another ancestor is determined through the same process as the first.  The entire algorithm is ran for $G$ generations, where $g=0$ initializes, and the remaining $G-1$ steps involve choosing ancestors and offspring.  For $g=0$, \texttt{InitializeKernel()} constructs a randomly initialized kernel, where the length is uniformly chosen from  $\{3, 5, 7, 9\}$ (respectively $2, 3, 4, 5$ base kernels), where the base kernels are uniformly sampled from $\mathcal{K}$, and add/multiply operations are equally likely. Experiments where we initialized with a more diverse set (kernel length from 1 to 15) slowed GA convergence with a negligible effect on end results. Our initialization was chosen with the goal of providing as unbiased of a sampling procedure as possible.  Alternative initializations could be used, e.g.~imposing a diversification constraint on the Age, Period, and Cohort terms in a given kernel, or infusing the initial generation with known-to-be-adequate kernels. It is also worth mentioning that although we fixed $n_g=200$ for all $g = 0, \ldots, G-1$, this value could vary over $g$ so that the total number of results is $\sum_{g=0}^{G-1} n_g$.

\begin{rem}
As mentioned, we utilized a double tournament method to combat bloat. Specifically in Algorithm \ref{algo:GA_algo}, whenever \texttt{DetermineAncestor} is run, it is in fact run twice to obtain $A^{g-1}_{i,(1)}$ and $A^{g-1}_{i,(2)}$, and the smaller length winner $A^{g-1}_i = A^{g-1}_{i,(j_0)}$ is chosen with probability $D/2 = 0.6$, where $j_0 = \arg\min_{j=1, 2} |A^{g-1}_{i,(j)}|$. See \cite{luke2006comparison} for details.  This is left out of Algorithm \ref{algo:GA_algo} for brevity.
\end{rem}

Since top-performing kernels are preferred as tournament winners, they become ancestors for future generators and are likely to re-appear as duplicates (either due to a \texttt{Copy} operation or a couple of mutations canceling each other). As a result, we observe many duplicates when aggregating all $\kappa_i^g$'s across generations.

\subsection{GP Hyperparameter Optimization}\label{sec:hyper_optim}

As mentioned, the hyperparameters of a given kernel $k(\cdot, \cdot; \theta)$ are estimated by maximizing $l_k(\theta| \mathbf{y})$, the marginal log likelihood of the observed data. The optimization landscape of kernel hyperparameters in a Gaussian process is non-convex with many local minima, so we take care in this optimization.  Since $\mathbf{y}$ is on the log scale, we leave these values untransformed.  For given $\bx = [x_{a}, x_{y}, x_{c}]^\top \in \R^3$, we perform dimension-wise scaling to the unit interval, e.g.~if $x_a = [x_{1,a}, \ldots, x_{n,a}]^\top$ then $x_{i,a} \mapsto \frac{x_{i,a} - \min(x_a)}{\max(x_a)-\min(x_a)}$ and similarly for $x_y$ and $x_c$.  
Lengthscales for stationary kernels can be interpreted on the original scale through the inverse transformation $\ell_{\text{len}} = \left(\max(x) - \min(x)\right) \tilde{\ell}_{\text{len}}$, where $\tilde{\ell}_{\text{len}}$ is on the transformed scale, with similar transformations for the mean function parameters.  For interpretability purposes, we utilize this to report values on the original scale in Sections \ref{sec:synthetic} and \ref{sec:hmd-results} whenever possible.  Non-stationary kernel (i.e.~$\Min, \Meh$ and $\Lin$) hyperparameters are left transformed, as they generally cannot be interpreted on the original scale.

We use Python with the GPyTorch library \cite{gardner2018gpytorch} to efficiently handle data and matrix operations.  Since our results rely heavily on accurate likelihood values (through BIC), we turn off GPyTorch's default matrix approximation methods.  Maximizing $l_k(\theta| \mathbf{y})$ is done using Adam \cite{kingma2014adam}.  This is an expensive procedure, as a naive evaluation of $l_k(\theta| \mathbf{y})$ is $\mathcal{O}(n^3)$ from inverting $\mathbf{K}(\mathbf{X},\mathbf{X})$.  Our GA runs use a convergence tolerance of $\varepsilon = 10^{-4}$ and maximum iterations of $\eta_{\text{max}} = 150$.  We found most simple kernels to converge quickly $(\eta \leq 100)$ even up to $\varepsilon = 10^{-6}$, see Table \ref{tab:conv_tol} in the Appendix.  We keep $\eta_{\text{max}}$ relatively small since we need to fit $n_g \cdot G \gg 10^3$ models.  Upon completion, the top few dozen kernels are refit with $\varepsilon = 10^{-6}$ and $\eta_{\text{max}} = 1000$.  Note that since Adam is a stochastic algorithm, the fitted kernel hyperparameters vary (slightly in our empirical work) during this refitting. This reflects the idea that the GA is a preliminary ``bird's-eye search" to find plausible mortality structures, thereafter refining hyperparameter estimates.  Except in a few cases, changes in final BIC values are minimal, though the ranking of top kernels can occasionally get adjusted.

\section{Synthetic Mortality Kernel Recovery}\label{sec:synthetic}

The premise of the GA is to carry out an extensive search that can correctly identify appropriate covariance structure(s) for a given dataset $\mathcal{D}$. The GA can be thought of as an initial search to yield a few thousand ($n_g \cdot G$) candidates, after which one can identify the top few best-fitting kernels as the ones to best represent the covariance structure of ${\cal D}$. To assess the quality of this kernel discovery process, we first try three synthetic datasets where the true data generating process, i.e.~the APC covariance structure, is given, and therefore we can directly compare the outputs of the GA to a known truth. This checks whether the GA can recover the true covariance, and by using Bayes factors, analyzes which kernels express similar information by substituting others.  Furthermore, it allows to assess the effect of noise on kernel recovery.  These experiments also provide a calibration to understand evidence categories for Bayes factors (see Table \ref{tab:bf_evidence_categories}) in the context of GP kernel comparison.  Simpler experiments focus on the restricted set of kernels $\cK_r$ as described in \ref{sec:kernel_details}, noting that $\Lin_y$ is the only appearance of linear (to model linear mortality improvement/decline in calendar year).  Additional experiments utilize the \emph{full set} $\cK_f$ of kernels, which includes all kernels in Table \ref{tab:kernels} over all coordinate dimensions.

The full experimental setup is  as follows.  First, we fix a GP kernel in the Age, Period, Cohort space and generate a respective log-mortality surface by sampling exactly from that prior. This creates a synthetic mortality surface. Below our surface spans Ages 50--84 and Years 1990--2019, with a linear parametric trend $\beta_0 + \beta_{ag} x_{ag}$ in Age. Thus, our training sets consist of $35 \cdot 30 = 1050$ inputs, identical in size to the HMD datasets used in Section \ref{sec:hmd-results}.

A three letter code is used to identify each synthetic mortality structure.  The first two experiments (SYA, SYB) use identically distributed, independent observation noise $\epsilon \sim {\cal N}(0, \sigma^2)$, and the third (SYC) takes heteroskedastic noise $\epsilon(\mathbf{x}) \sim \mathcal{N}(0, \sigma^2 / D_\mathbf{x})$, with $D_{\mathbf{x}}$ coming from the real-world HMD Japan Female dataset to capture realistic heterogeneity in age.  The precise setups along with the hyperparameters are described in Table \ref{tab:synthetic-description}, and the resulting synthetic mortality surfaces are available for public re-use at \url{github.com/jimmyrisk/gpga-synthetic-surfaces}.  All of the synthetic kernels and their hyperparameters are real-world-plausible. Namely SYA, SYB, SYC all came from prototypical GA runs on HMD datasets and hence match their structure.

\begin{table}
    \centering
    \begin{tabular}{c|p{3.4in}|lll} \toprule
        Exprmnt & Ground Truth Kernel & $\sigma^2(\mathbf{x})$ & $\beta_0$ & $ \beta_{ag}$ \\ \midrule
        SYA & $0.04 \cdot \RBF_a(0.4) \cdot \RBF_y(0.3)$ & 0.001 & -5.0 & 3.4\\
        SYB & $0.08 \cdot \RBF_a(0.586) \cdot M12_y(13.33) + 0.02 \cdot M52_c(0.079)$ & 0.0004 & -5.568 & 2.974\\
        SYC & $0.0134 \cdot M52_a(1.132) \cdot \Min_y(0.877) \cdot M12_c(96.234)\cdot$ & $1.0783 / D_\mathbf{x}$ & -3.165 & 3.380\\
         & $ \qquad \cdot \Meh_c(0.8483) $ &  & \\
        \bottomrule
    \end{tabular}
    \caption{Description of synthetic data sets.  Data is generated as multivariate normal realizations according to the Equations in \eqref{eq:y-mean-var}, with parametric mean function $m(\mathbf{x}) = \beta_0 + \beta_{ag} x_{ag}$.  SYA and SYB are homoskedastic.  In generating SYC's heteroskedastic noise, $D_\mathbf{x}$ comes from the JPN Female data; see Section \ref{sec:hmd-results} for details regarding this data set.}
    \label{tab:synthetic-description}
\end{table}

The first case study (SYA) starts with an exceptionally simple structure (kernel of length 3, i.e.~2 base kernels) as a check on whether the GA preserves parsimony when exploring the kernel space. In other words, we use SYA to validate that longer-length kernels would correctly be perceived as over-fitting during the GA evolution, and receive lower fitness scores compared to the true kernel. As a secondary effect, SYA addresses recovery of kernel smoothness, as $\RBF_a \cdot \RBF_y$ is smooth in both age and year components.  Since we minimize BIC for the given training set, it is plausible that a different kernel from the generating one might actually achieve a (slightly) lower BIC for a given realization, so this experiment is performed twice (generating SYA-1 and SYA-2) to assess  sampling variability.

The second synthetic example (SYB) features a more sophisticated kernel of length 7 (with 3 base kernels), and moreover combines both multiplicative and additive structure. It is motivated by Lee-Carter models and has a multiplicative age-period component with a less prominent additive cohort effect (coefficient of 0.02 versus 0.08). Its purpose is to (i) check whether the GA is able to identify such high-level structure (including addition and multiplication, on the correct terms), (ii) identify low-length kernels as under-fitting, and (iii) distinguish between all three of age, year, and cohort terms.

The third (SYC) is purely multiplicative with a more complicated kernel period-cohort structure with 4 components representative of results from real world experiments. The Min${}_y$ kernel adds a nonstationary period effect, and there is a sophisticated cohort effect consisting of a smooth non-stationary Mehler term, as well as the rough $\M12_c$. Note that the large lengthscale $\ell_{\text{len}}=6062.76$ corresponds to a nearly non-stationary process. By imposing a separable structure with four kernels in the three-dimensional input space, this experiment investigates GA's ability to recover nonstationarity and roughness across multiple dimensions in the presence of possible over-parameterization and/or collinearity. Furthermore, since $\Meh \notin \cK_r$, we use SYC as an experiment to assess the GP's ability to approximate the $\Meh$ covariance structure through simpler kernels in $\cK_r$ by performing a GA run both with $\cK_r$ and with $\cK_f$.

Both SYA and SYB are done purely searching over $\cK_r$, with SYA testing a basic APC setup, and SYB emphasizing recovering additive structure and linear coefficients.  Using $\cK_f$ could muddle the analysis, and is therefore left for SYC. Random number generator seeding for the initial generation is unique to each GA run; all GA runs have identical hyperparameters as described \ref{tab:ga_params_high_level} and \ref{tab:ga_params_high_level_2}.

\begin{rem}
The theoretical Bayes factor in Equation \eqref{eq:bayes_factor_bf} assumes all hyperparameters have been integrated out $p(y|k) = \int_\theta p(y|k,\theta)d\theta$, or replaced with MLE's when using BIC.  Thus, the hypothesis being tested through the Bayes factors is purely about the kernel choice.
\end{rem}

\subsection{Synthetic Results}

Answers to questions presented in the previous section are found in Tables \ref{tab:SYA-results} for SYA, Table \ref{tab:SYB-results} for SYB, and Table \ref{tab:SYC-results} for SYC.   In all tables, $K_0$ denotes the known kernel that generated the synthetic data.  With the goal in mind to establish kernel recovery (ignoring hyperparameter estimation), $\hat{\theta}$ is estimated for $K_0$ from the generated data.

\subsubsection*{SYA}

\begin{table}[t]
    \centering
    \begin{tabular}{cll|cll}\toprule
     \multicolumn{3}{c|}{SYA-1} & \multicolumn{3}{|c}{SYA-2} \\ \midrule
     BIC & \multicolumn{1}{c}{$\widehat{\text{BF}}(k, K_0)$} & Kernel & BIC & \multicolumn{1}{c}{$\widehat{\text{BF}}(k, K_0)$}  & Kernel \\
    \textbf{-2034.23} & \textbf{1.0000***} & $\mathbf{\textbf{RBF}_a \textbf{RBF}_y}$ & -2066.93 & 1.1907*** & $\M52_a \RBF_y$ \\
    -2034.04 & 0.8264*** & $\M52_a \RBF_y$ & \textbf{-2066.76} & \textbf{1.0000***} & $\mathbf{\textbf{RBF}_y  \textbf{RBF}_a}$  \\
    -2031.82 & 0.0902* & $\M52_a \M52_y$ & -2064.63 & 0.1216** & $\M52_a \M52_a\RBF_y  $ \\
    -2031.29 & 0.0526* & $\M52_a  \RBF_a \RBF_y$ & -2064.24 & 0.0801* & $ \M52_a   \RBF_a \RBF_y$ \\
    -2031.09 & 0.0433* & $\M52_a  \M52_a \RBF_y$ & -2063.88 & 0.0561* & $  \M52_a  \M52_a \RBF_y$   \\ \bottomrule
    \end{tabular}
    \caption{Top five fittest non-duplicate kernels for the first synthetic case study SYA.  Bolded is $K_0 = \RBF_y  \RBF_a$, the true kernel used in data generation.  SYA-1 and -2 denote the realization trained on.}
    \label{tab:SYA-results}
\end{table}

Table \ref{tab:SYA-results} shows the results of the top 5 fittest kernels for SYA-1 and SYA-2.  The estimated Bayes factor (using BIC) is the column $\widehat{\text{BF}}(k, K_0) = \exp(\BIC(K_0) - \BIC(k))$.  For SYA-1, the true kernel $K_0 = \RBF_y  \RBF_a$ is discovered and appears with lowest BIC.  Next best (with a large Bayes factor of $0.826$)  is $\M52_a  \RBF_y$, which has an identical structure aside from using the slightly less smooth Mat\'ern-$5/2$ kernel for age instead of RBF.  Note that the Bayes factor would need to be below $1/3 \approx 0.3333$ to even be worth mentioning a difference between these kernels according to Table \ref{tab:bf_evidence_categories} in the appendix, suggesting an indifference of the two models.  Our interpretation is that sampling variability makes $M52_a$ a similar alternative to $RBF_a$.  Note, however, that any lower Mat\'ern order does not appear in either table.  The remaining three rows have a similar kernel structure with superfluous Age or Period kernels added.  The Bayes factors are below 0.1,  suggesting  strong evidence against these kernels as being plausible alternatives for the generated dataset.  This experiment confirms the GA's ability to recover the overall structure, with an understandable difficulty in distinguishing between $M52_a$ and $\text{RBF}_a$.  SYA-2 in the right panel of Table \ref{tab:SYA-results} considers a different realized mortality surface under the same population distribution to assess stability across GA runs. It tells a similar story, though interestingly the lowest-BIC kernel is now $\M52_a \RBF_y$, with $\widehat{BF}(k, K_0) = 1.1907$. This means that it achieves a lower BIC than the true kernel, showcasing possibility of overfitting to data. At the same time, since the BF is so close to 1, this is still not statistically worth mentioning and hence can be fully chalked up to sampling variability. Otherwise we again observe only two truly plausible alternatives, and very similar less-plausible (BF between 0.05 and 0.13) alternates.

To further assess smoothness detection, multiplicative Age-Period kernels $k=k_a k_y$ are fit to both SA1 and SA2 datasets, where $k_a$ and $k_y$ range over $\M12, \M32, \M52$ and $\RBF$ (in order of increasing smoothness), resulting in a total of 16 combinations per training surface.  The resulting differences in BIC are provided in Table \ref{tab:SA_smoothness}.  Note that the ground truth kernel $K_0 = \RBF_a \RBF_y$  generates a mortality surface that is infinitely differentiable  in both Age and Period. In both cases, decisive evidence is always against either surface having a $\M12$ component (with $\BIC(K_0) - \BIC(k)$ ranging from $-96.99$ to $-30.24$ -- recall that anything below -4.61 is decisive evidence for $K_0$).  As found above, $\M52_a$ is a reasonable surrogate for $\RBF_a$ for both SYA-1 and SYA-2.  SYA-1 shows only strong evidence against $\M52_y$ (as opposed to decisive for SYA-2).  In both cases, there is decisive evidence against $\M32_y$, with only strong evidence on the age component $\M32_a$ when using $\RBF_y$ (otherwise, it is decisive).  This difference is likely explained by the additional flexibility of $\M32_a$ to pick up some fluctuations which would normally be reserved for the parametric mean function in age.

\begin{table}[]
    \centering
\begin{tabular}{l|llll|llll} \toprule
& \multicolumn{4}{c|}{SYA-1} & \multicolumn{4}{|c}{SYA-2}\\
\midrule
{} &  $\M12_a$ &  $\M32_a$ &  $\M52_a$ &  $\RBF_a$&  $\M12_a$ &  $\M32_a$ &  $\M52_a$ &  $\RBF_a$ \\
\midrule
$\M12_y$ & -63.98 & -38.96 & -37.49 & -39.41 & -96.99 & -68.85 & -66.12 & -65.25 \\
$\M32_y$ & -35.83 &  -8.13 &  -5.98 &  -7.12 & -53.24 & -21.19 & -17.67 & -16.06 \\
$\M52_y$ & -32.29 &  -4.70 &  -2.41* &  -3.33* & -42.75 & -11.36 &  -8.09 &  -6.92 \\
$\RBF_y$ & -30.24 &  -3.14* &  -0.19*** &   0.00*** & -32.67 &  -2.31* &   0.17*** &   0.00*** \\
\bottomrule
\end{tabular}
    \caption{Logarithmic Bayes Factor $\log\widehat{BF}(k, K_0) = \text{BIC}(K_0) - \text{BIC}(k)$ for $K_0 = \RBF_a \RBF_y$ and $k = k_a k_y$, where $k_a$ and $k_y$ are in the respective row and column labels and $\text{BIC}(K_0)$ is evaluated over SYA-1 (left panel) and SYA-2 (right panel).}
    \label{tab:SA_smoothness}
\end{table}

\subsubsection*{SYB}

\begin{table}[t]
    \centering\footnotesize
    \begin{tabular}{ccl|ccl} \toprule
     \multicolumn{6}{c}{SYB}  \\ \midrule
     BIC & $\widehat{\text{BF}}(k, K_0)$ & Kernel & $\hat{\sigma}^2$ & $\hat{\beta}_0$ & $\hat{\beta}_{ag}$ \\
     -2468.0 & 1.033*** & $0.014\cdot \RBF_a(0.574)   \Min_y(4.4) +0.018 \cdot \M52_c(0.083)$ & $3.52 \cdot 10^{-4}$ & $-4.249$ & $3.264$\\
     \textbf{-2467.8} & \textbf{1.000***} & $\mathbf{0.063 \cdot \textbf{RBF}_a(0.585)  \textbf{M12}_y(8.937)+0.0180\cdot \textbf{M52}_c(0.084)}$ & $\mathbf{3.448 \cdot 10^{-4}}$ & $\mathbf{-4.145}$ & $\mathbf{3.332}$\\
     -2465.9 & 0.161** & $0.060\cdot \RBF_a(0.574)  \M12_y(8.267)+0.016\cdot \RBF_c(0.051)$ & $3.60 \cdot 10^{-4}$ & $-4.249$ & $3.264$\\
     -2465.7 & 0.127** & $0.014\cdot \RBF_a(0.574) \Min_y(4.3) +0.016\cdot \RBF_c(0.051)$ & $3.60 \cdot 10^{-4}$ & $-12.01$ & $3.264$\\
     -2464.7 & 0.047* & $0.022\cdot \RBF_a(0.591)  \Lin_y(2.8) \M12_y(10.480)+0.016\cdot \M52_c(0.083)$ & $3.45 \cdot 10^{-4}$ & $-4.312$ & $3.298$\\
     \bottomrule
    \end{tabular}
    \caption{Top five fittest non-duplicate kernels for the second synthetic case study SYB.  Bolded is the true kernel used in data generation $K_0 = 0.08 \cdot \RBF_a(0.586) \cdot M12_y(13.33) + 0.02 \cdot M52_c(0.079)$, with $\sigma^2 = 4 \cdot 10^{-4}, \beta_0 = -5.568 , \beta_1 = 2.974$.  }
    \label{tab:SYB-results}
\end{table}

Next we discuss SYB where the training surface is generated from $K_0 = 0.08 \cdot\RBF_a \M12_y + 0.02 \cdot \M52_c$.  Table \ref{tab:SYB-results} shows that there a total of four plausible kernels found (with Bayes Factors above 0.1). The GA discovers the ground truth (bolded in Table \ref{tab:SYB-results}) with slightly different hyperparameters, and three closely related alternates. In particular, these four fittest kernels all identify the correct number of APC terms with a multiplicative Age-Period term plus an additive Cohort term. The $\RBF_a$ term is recovered in all four, and the corresponding Age lengthscales are very close to the true $\ell^a_{\text{len}} = 0.586$. The Cohort effect is captured either via the ground truth $\M52_c$ (top-2 candidates) or $\RBF_c$, a substitution phenomenon similar to what we observe for SYA above. The respective lengthscale is correctly estimated to be within 0.3 of the true $\ell^c_{\text{len}}=0.079$. The Period effect is captured either by the ground truth $\M12_y$ or by $\Min_y$. The data-generating $K_0$ has a very large $\ell^y_{\text{len}} = 13.33$ for the $\M12_y$ term
which corresponds to an AR(1) process with $\phi_0=0.9975$ after unstandardizing then transforming. Over the training range of 30 years this is almost indistinguishable from a random walk with a $\Min$ kernel. As a result, the substitution with $\Min_y$ is unsurprising, as is the wide range of estimated $\ell_{\text{len}}$. For example, $\hat{\ell}^y_{\text{len}}=8.933$ in the second row corresponds to AR(1) persistence parameter of $\phi = 0.9960$, very close to the true $\phi_0$. We furthermore observe stable recovery of all non-kernel hyperparameters $(\sigma^2, \beta_0, \beta_{ag})$, with estimates close to their true values.

Finally, the GA is also very successful in learning that the Age-Period component (coefficient 0.08) dominates the Cohort component (coefficient 0.02), conserving the relative amplitude of the two components.
Note that the $\Min$ and $\Lin$ kernels include an offset which mathematically result in similar linear coefficients to the true 0.08 and 0.02.  For example, the first term in the first row simplifies to $0.0141 \cdot \RBF_a \cdot (4.42 + x_{yr} \wedge x'_{yr}) = 0.0623 \cdot \RBF_a + 0.0141 \cdot \RBF_a \cdot (x_{yr} \wedge x'_{yr})$.  This results in all five kernels in Table \ref{tab:SYB-results} discovering the first component to be larger than the second by a factor of 3.5--4.

\subsubsection*{SYC}

\begin{table}[t]
    \centering
    \begin{tabular}{ccl|ccl} \toprule
         \multicolumn{6}{c}{SYC} \\
    \multicolumn{3}{c|}{Restricted search set $\cK_r$} & \multicolumn{3}{c}{Full search set $\cK_f$}\\ \midrule
       BIC & $\widehat{\text{BF}}(k, K_0)$ & Kernel & BIC & $\widehat{\text{BF}}(k, K_0)$  & Kernel \\
-2785.35 & 0.1930** & $\M52_a \Min_y \M12_c \RBF_c$ & -2786.82 &  0.8321*** &  $\textbf{M52}_a \M12_y \textbf{M12}_c \textbf{Meh}_c$ \\
-2785.06 & 0.1444** & $\M52_a \Min_y \M12_c \M52_c$ & -2785.49 &  0.2218** &  $\M52_a \M12_y \Min_c \RBF_c$  \\
-2784.99 & 0.1339** & $\M52_a \M12_y \M12_c \RBF_c$ & -2785.41 &  0.2046** &  $\Chy_a \Min_y \M12_c \Meh_c$  \\
-2784.73 & 0.1032** & $\M52_a \M12_y \M12_c \M52_c$ & -2785.21 &  0.1671** &  $\Chy_a \M12_y \M12_c \Meh_c$ \\
-2784.21 & 0.0615* & $\M52_a \Min_y \RBF_y \M12_c$  & -2784.87 &  0.1185** &  $\Chy_a \M12_y \Meh_y \M12_c$ \\ \bottomrule
    \end{tabular}
    \caption{Fittest non-duplicate kernels for SYC in two separate runs, one over $\cK_r$ and the other over $\cK_f$.  The true kernel is $K_0 = \M52_a \Min_y (\M12_c \Meh_c)$. Note that $\Meh_c \notin \cK_r$.}
    \label{tab:SYC-results}
\end{table}

Table \ref{tab:SYC-results} shows the results for SYC.  The left panel (results over $\mathcal{K}_r$) shows GP's ability to approximate $K_0 = \M52_a \Min_y  (\M12_c \Meh_c)$ through other base kernels given that $\Meh_c \notin \cK_r$.  Multiple features of $K_0$ are recovered perfectly: the Age component, with $\M52_a$ being the only Age term appearing; the multiplicative structure with correct number of kernels (four), and the presence of $\M12_c$ for the Cohort effect.  The $\Meh_c$ term is replaced either with $\RBF_c$ or with $\M52_c$. Note that the best found kernel in $\mathcal{K}_r$ has a Bayes factor of only $\widehat{BF}(k, K_0)=0.1930$, illustrating that this substitution is imperfect and that the type of nonstationarity that $\Meh$ adds over $\RBF$ is statistically significant.

Looking at the right panel, there is a clear winner in $\M52_a \M12_y \M12_c \M52_c$ which only mis-identifies $\Min_y$ rather than $\M12_y$, a similar substitution as observed for SYB. This kernel has a BF above 0.83 and  according to the Bayesian paradigm, is $3.75 = 0.8321 / 0.2218$ times more likely for the training data compared to the next choice in the second row (switching $\M12_c$ to $\Min_c$, and $\Meh_c$ to $\RBF_c$).  As in the left panel, this helps to establish the prominence of $\Meh$'s nonstationarity.  The results are otherwise similar to $\cK_r$ with an inclusion of $\Chy_a$ to substitute $\M52_a$. Interestingly, the ground truth $K_0$ is dodged in this run of the GA, even though it could have been obtained through a single cross-over or non-subtree mutation from other top-performing kernels. This suggests that the GA was not able to fully explore the kernel space in 20 generations, likely influenced by the slow  convergence in hyperparameter estimation when $\Min$ is present.  For further discussion, see Appendix \ref{sec:appendix_convergence}.

The SYC synthetic surface also illustrates the GA's ability to identify the number of total terms in $K_0$ and the presence of an additive structure. Using the shorthand $\widehat{\BF} = \widehat{\BF}(k, K_0)$ the GA finds that:

\begin{itemize}
    \item The best kernel with three components (rather than four) is $\Chy_a \M12_y \M12_c$ ($\widehat{\BF} = 0.0206$) in $\mathcal{K}_f$ and $\M52_a \M12_y \M12_c$ ($\widehat{\BF} = 0.0182$) in $\mathcal{K}_r$. Thus, the BIC criterion leads the GA to correctly reject such kernels as being too short for the SYC dataset;
    \item The best kernel with five components is $\M52_a (\M12_y \M12_y)(\M12_c \Meh_c)$ ($\widehat{\BF} = 0.0108$) in $\mathcal{K}_f$ and $\M52_a \M12_y \Lin_y \M12_c \M52_c$ ($\widehat{\BF} = 0.0108$) in $\mathcal{K}_r$. Again, the GA and its BIC criterion  correctly penalizes kernels that are too long.
    \item The best kernel with an additive component is $\M52_a (\M12_y + \M12_c \RBF_c)$ ($\widehat{\BF} = 0.0070$) and $\Chy_a \Meh_c (\M12_c + \M12_y)$ ($\widehat{\BF} = 0.0067$) in $\mathcal{K}_r$. This is decisive evidence that the BIC properly learns the presence of a single additive term.
\end{itemize}

\section{Results on Human Mortality Database data} \label{sec:hmd-results}

After validating our generative kernel exploration with synthetic data, we move to realistic empirical analysis.
Our discussion focuses on retrospective analysis, namely the performance of different kernels assessed in terms of fitting a given training set. Thus, we do not pursue out-of-sample metrics, such as (probabilistic) scores for predictive accuracy and concentrate on looking at the BIC scores augmented with a qualitative comparison. Retrospective assessment parallels the core of APC methods that seek to decompose the data matrix via Singular Value Decomposition, prior to introducing the out-of-sample dynamics in the second step. A further reason for this choice is that predictive accuracy is fraught with challenges (such as handling idiosyncratic data like the recent 2020 or 2021 mortality driven by COVID), and there is no canonical way to assess it. In contrast, BIC offers a single ``clean" measure of statistical fit for a GP and moreover connects to the BF interpretation of relative preponderance of evidence.

As our first case study we consider the Japanese Female population. We utilize the HMD dataset covering Ages 50-84 and years 1990-2018. The same top-level and crossover/mutation hyperparameters of the GA are used as in the last section (Tables \ref{tab:ga_params_high_level} and \ref{tab:ga_params_high_level_2} respectively).  

Table \ref{tbl:jpn-robust} provides summary statistics regarding the top kernels in ${\cal K}_f$ that achieve the lowest BIC scores. In order to provide a representative cross-sectional summary, we consider statistics for the top-10, top-50, and then 51-100, 101-150 and 151-200th best kernels. Recall that there are a total of $20 \cdot 200 = 4000$ kernels proposed by the GA, so top-200 correspond to the best 5\% of compositions. We find that the best fit is provided by purely multiplicative kernels (single additive component) with 3 or 4 terms. This includes one term for each of Age, Period and Cohort coordinates, plus a possible 4th term, usually in Cohort or Year. In the restricted class ${\cal K}_r$, all of the top-10 kernels are of this form, as are 9 out of top-10 kernels found in ${\cal K}_f$.

\begin{table}[h!]
\begin{tabular}{lrrrrrrrrrrr}
\toprule
Range &  BIC &  BIC &  len &  addtv  &  non- &  num & num & num & rough  &  rough &  rough   \\
  & max & min &  & comps & stat. & age & year & coh & age & year & coh \\
   \midrule  \multicolumn{12}{c}{JPN Female} \\ \midrule
 1-10 & -2723.68 & -2725.29 & 4.00 & 1.00 & 0\% & 1.00 & 1.80 & 1.20 & 0\% & 100\% & 100\% \\
 1-50 & -2720.64 & -2725.29 & 4.34 & 1.08 & 10\% & 1.12 & 1.90 & 1.32 & 0\% & 100\% & 100\% \\
 51-100 & -2718.24 & -2720.62 & 4.60 & 1.20 & 18\% & 1.12 & 2.20 & 1.28 & 0\% & 100\% & 100\% \\
 101-150 & -2717.03 & -2718.17 & 5.02 & 1.14 & 4\% & 1.30 & 2.18 & 1.54 & 6\% & 98\% & 100\% \\
 151-200 & -2715.77 & -2717.01 & 5.10 & 1.48 & 12\% & 1.28 & 2.36 & 1.46 & 6\% & 100\% & 100\% \\
   \midrule  \multicolumn{12}{c}{JPN Female Rerun} \\ \midrule
 1-10 & -2723.05 & -2725.29 & 4.00 & 1.00 & 0\% & 1.00 & 1.60 & 1.40 & 0\% & 100\% & 100\% \\
 1-50 & -2719.82 & -2725.29 & 4.14 & 1.08 & 10\% & 1.08 & 1.58 & 1.48 & 0\% & 98\% & 100\% \\
 51-100 & -2718.30 & -2719.82 & 4.46 & 1.26 & 8\% & 1.14 & 1.62 & 1.70 & 6\% & 100\% & 100\% \\
   \midrule  \multicolumn{12}{c}{JPN Female Search in ${\cal K}_r$} \\ \midrule
 1-10 & -2724.11 & -2725.27 & 4.00 & 1.00 & 0\% & 1.00 & 1.70 & 1.30 & 0\% & 100\% & 100\% \\
 1-50 & -2721.19 & -2725.27 & 4.48 & 1.10 & 8\% & 1.14 & 1.96 & 1.38 & 0\% & 100\% & 100\% \\
 51-100 & -2718.06 & -2721.19 & 4.72 & 1.50 & 18\% & 1.16 & 1.96 & 1.60 & 0\% & 100\% & 100\% \\
  \midrule  \multicolumn{12}{c}{JPN Female trained on ${\cal D}_{rob}$} \\ \midrule
1-10 & -2724.11 & -2725.29 & 4.00 & 1.40 & 40\% & 1.00 & 1.50 & 1.50 & 0\% & 100\% & 100\% \\
1-50 & -2716.84 & -2725.29 & 4.42 & 1.12 & 18\% & 1.14 & 1.64 & 1.64 & 0\% & 100\% & 100\% \\
 51-100 & -2714.96 & -2716.58 & 4.70 & 1.16 & 12\% & 1.18 & 1.68 & 1.84 & 0\% & 100\% & 100\% \\
\bottomrule
\end{tabular}
\caption{Re-run and robust check for JPN Females across both ${\cal K}_r$ and ${\cal K}_f$ \label{tbl:jpn-robust}}
\end{table}

The above APC structure moreover includes a rough (non-differentiable or only once-differentiable) component in Year and in Cohort. This matches the logic of time-series models for evolution of mortality over time. Note that in our setting, it can be interpreted as a strong correlation of observed noise across Ages, in other words the presence of environmental disturbances (epidemics, heat waves, other co-morbidity factors) that yield year-over-year idiosyncratic impacts on mortality.  On the other hand, in Age best fits are smooth, most commonly via the M52 kernel. This matches the intuition that the Age-structure of mortality is a smooth function. Table \ref{tbl:jpn-robust} documents a strong and unequivocal cohort effect: present 100\% in all top kernels. This is consistent with \citet{willets2004cohort} who states that a strong cohort effect for the Japanese Female population can be projected into older ages.

The presence of multiple factors (length above 3) generally indicates one or both of the following: (i) multi-scale dependence structure; (ii) model mis-specification. On the one hand, since we are considering only a few kernel families, if the true correlation structure is not matched by any of them, the algorithm is going to substitute with a combination of the available kernels. Thus, for example using both a rough and a smooth kernel in Year indicates that neither of the M12 or RBF fit well on their own. On the other hand, the presence of additive structure, or in general the need for many terms (especially over 5) suggests that there are many features in the correlation structure of the data, and hence it does not admit any simple description.

In JPN Females, the GA's preference for parsimony is confirmed by the fact that the best performing kernels are the shortest. We observe a general pattern that Length is increasing in Rank. In particular, going down the rankings, we start to see kernels with two additive components. We may conclude that the second additive component provides a minor improvement in fit, which is outweighed by the complexity penalty and hence rejects on the grounds of parsimony. 

The compositional kernel that achieves the lowest overall BIC is $$k^*_{JPN-FEM} = 0.4638 \cdot M52_a(1.11) \cdot  \Chy_y(1.95)\cdot M12_y(62.42)\cdot  M12_c(117.11).$$
Note the purely multiplicative structure of $k^*_{JPN-FEM}$ and its two Period terms, capturing both the local rough nature and the longer-range dependence. Table \ref{tab:JPN-best-results} lists the next-best alternatives, both within ${\cal K}_f$ and within ${\cal K}_r$. We see minimal loss from restricting to the smaller ${\cal K}_r$, as three of five top kernels found in ${\cal K}_f$ actually belong to ${\cal K}_r$. Thus, casting a ``wider net'' does not improve BIC, suggesting that most of the kernel options added to ${\cal K}_f$ are either close substitutes to the base ones in ${\cal K}_r$ or do not specifically help with HMD data. Indeed, the BF improvement factor is just $\exp(0.19)$ from Table \ref{tbl:jpn-robust}. Moreover, we also find that there is strong hyperparameter stability across different top kernels. For example, we find that the lengthscale in Age (which is always captured via a $\M52$ kernel in Table \ref{tab:JPN-best-results}) is consistently in the range $[1.09, 1.13]$. Similarly, the lengthscales for the $\M12$ kernel in Cohort are in the tight range $[95,120]$.

\begin{table}[ht!]
    \centering
    \begin{tabular}{rcl|rcl} \toprule
         \multicolumn{6}{c}{Japan Female HMD Dataset for 1990-2018 and Ages 50-84} \\
    \multicolumn{3}{c|}{$\cK_r$} & \multicolumn{3}{c}{$\cK_f$}\\ \midrule
       BIC & $\widehat{\text{BF}}$ & Kernel & BIC & $\widehat{\text{BF}}$  & Kernel \\
 -2725.288 & 0.995 & $\M52_a (\RBF_y \M12_y) \M12_c$ & -2725.293 &  1 &  $\M52_a (\Chy_y \M12_y) \M12_c$ \\
 -2725.270 & 0.977 & $\M52_a (\M52_y \M12_y) \M12_c$    & -2725.270  &  0.977$^\dag$ &  $  \M52_a (\M52_y \M12_y) \M12_c$ \\
 -2725.233 & 0.941 & $\M52_a (\RBF_y \Min_y) \M12_c$  & -2725.221 &  0.931$^\dag$ &  $ \M52_a (\M52_y \Min_y) \M12_c  $  \\
 -2725.221 & 0.931 & $\M52_a (\M52_y \Min_y) \M12_c$ & -2724.623 &  0.512$^\dag$ &  $ \M52_a (\M52_y \M12_y) \Min_c   $  \\
 -2724.640 & 0.520 & $\M52_a (\M52_y \M12_y) \Min_c$ & -2724.510 &  0.457 &  $\M52_a (\M32_y \M12_c) \M12_c$ \\
 \bottomrule
    \end{tabular}
    \caption{Fittest non-duplicate kernels for Japanese Females in two separate runs, one over $\cK_f$ and the other over $\cK_r$. Bayes Factors $\widehat{\text{BF}}$ are relative to the best found kernel $k^*_{JPN-FEM}$ and all have insubstantial significance. Daggered kernels under ${\cal K}_f$ column are those that also belong to ${\cal K}_r$ }
    \label{tab:JPN-best-results}
\end{table}

\begin{figure}[ht]
\begin{center}
\begin{tabular}{cc}
\includegraphics[height=2.2in,trim=0.65in 0.3in 0.45in 0.1in]{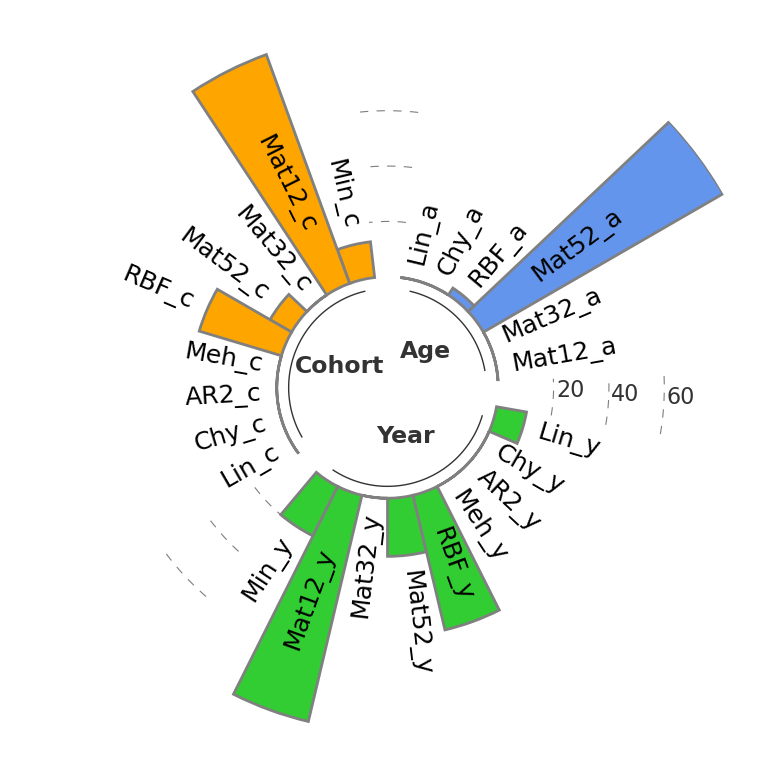} & \includegraphics[height=2.2in,trim=0.5in 0.6in 0.7in 0.1in]{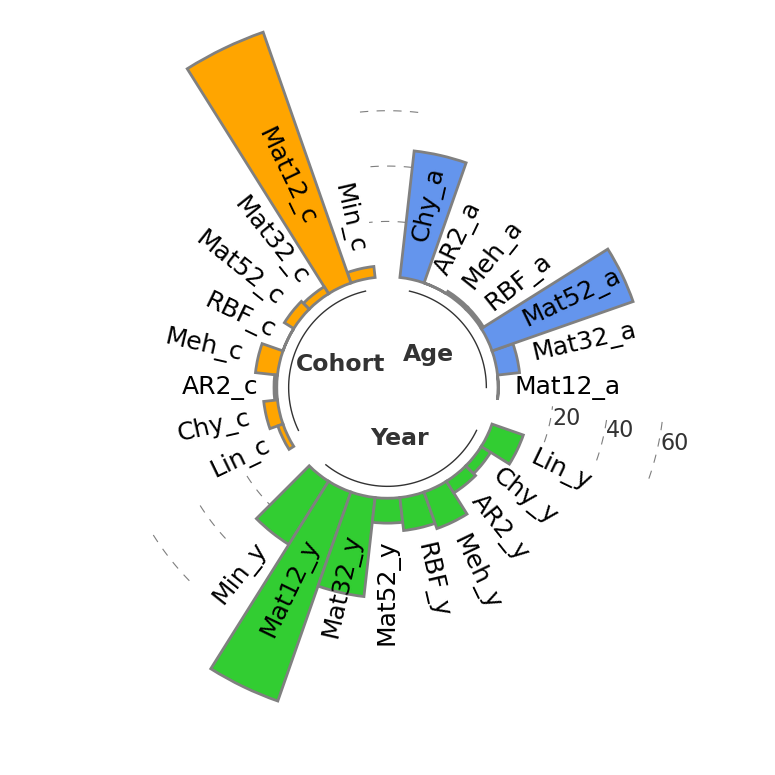} \\
${\cal K}_r$ & ${\cal K}_f$ \\
\end{tabular}
\end{center} \vspace*{-12pt}
\caption{Frequency of appearance of different kernels in JPN Female models. \label{fig:circBar-jpn}}
\end{figure}

Figure \ref{fig:circBar-jpn} visualizes the frequency of the appearance of different kernels. We consider the top 100 unique kernels returned by the GA and show the number of times each displayed kernel is part of the composite kernel returned. Note that sometimes the same kernel can show more than once. In the left panel we consider GA searching in ${\cal K}_r$, hence many of the families are not considered; the right panel looks at the full ${\cal K}_f$.

The barplots in Figure \ref{fig:circBar-jpn} indicate that there is a substantial variability in selected kernels when considering the top 100 proposals. Nevertheless, we observe the typical decomposition into ``principal'' factors, such as $\M12_c$ and $\M12_y$ for JPN Females, plus additional residual kernels. The latter generate second-order effects and are not easily identifiable, leading to a variety of kernels showing up for 5-15\% of the proposals. For example, nearly every kernel family in Period can be used to construct a good compositional kernel. This heterogeneity of kernels picked indicates that it is not appropriate to talk about ``the" GP model for a given dataset, as there are several,  quite diverse fits that work well.

\begin{figure}[!ht]
\begin{tabular}{cc}
\includegraphics[height=2.25in]{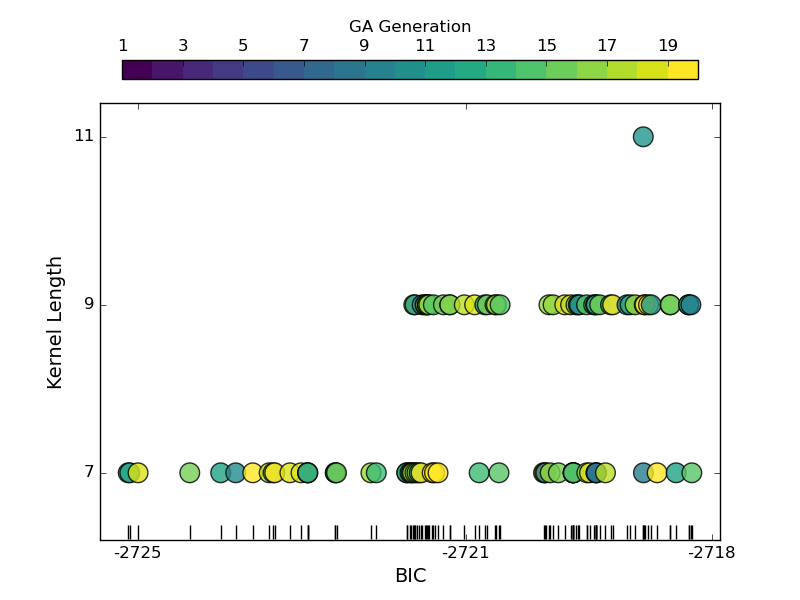} &
\includegraphics[height=2.25in]{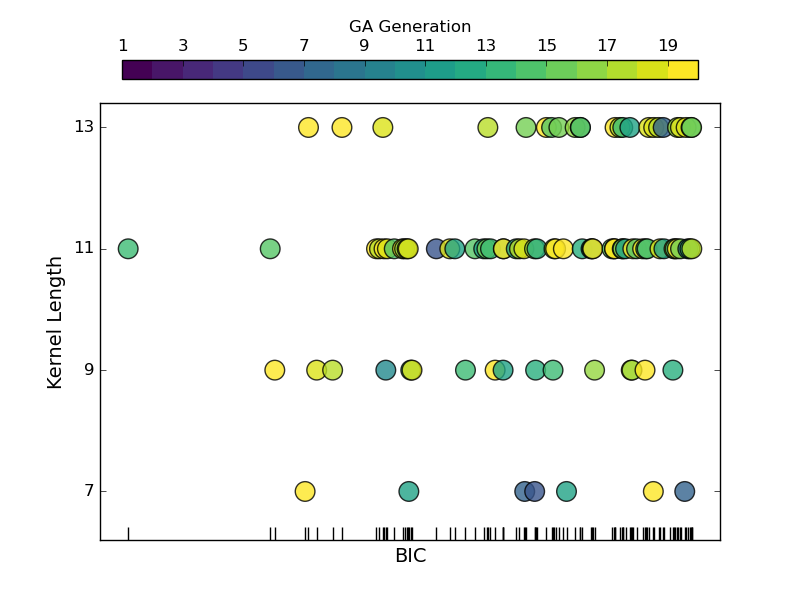} \\
JPN Female & USA Male \end{tabular}
\caption{Properties of the top 100 kernels found by GA. \label{fig:kernel-length}}
\end{figure}

Figure \ref{fig:kernel-length} shows several summary statistics of proposed kernels against their BIC scores. We display the top 100 unique kernels, arranged according to their total length (y-axis) and generation found (color). First, we observe that there is an increasing clustering of kernels as we march down the BIC order (x-axis). In other words, there is typically a handful (1-5) of best-performing kernels, and then more and more equally-good alternatives as the BIC decreases. This matches the interpretation of Bayes factors: accepting the best performing kernel as the ``truth", we find several plausible alternatives, a couple dozen of somewhat plausible ones, and many dozens of weakly plausible ones. The spread of the respective BF factors varies by population; in some cases there are only $\sim 50-60$ plausible alternatives, in others there are well over a hundred.

Second, we observe that most of the best kernels are found after 10+ generations, matching the logic of the GA exploring and gradually zooming into the most fit kernel families. However that pattern is not very strong, and occasionally the best kernel is discovered quite early on.

Third, we observe a pattern in terms of kernel complexity vis-a-vis its fitness, matching the above logic: the best performing kernels tend to be of same length (and are very similar to each other, often just 1 mutation away), but as we consider (weakly) plausible alternatives, we can find both more parsimonious and more complex kernels. This captures the parsimony trade-off: shorter kernels have lower likelihood but smaller complexity penalty; longer kernels have higher log-likelihood but are penalized more.

\begin{figure}
    \centering
    \includegraphics[width=0.44\textwidth]{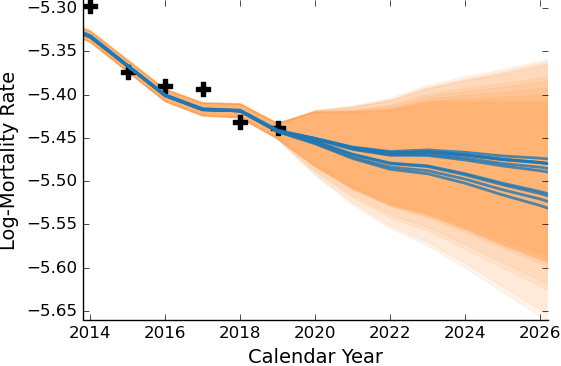}
    \includegraphics[width=0.4\textwidth]{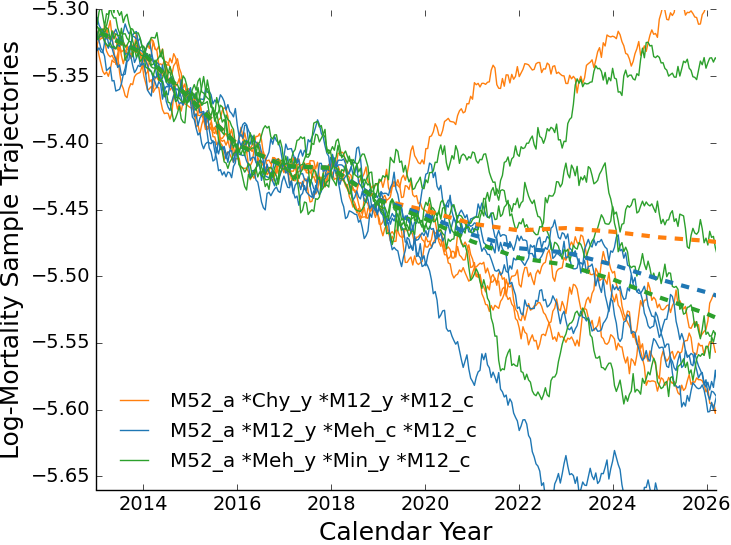}
    \caption{Predictions from the top 10 kernels in ${\cal K}_f$ for JPN Females Age 65. \emph{Left:} predictive mean and 90\% posterior interval from the top-10 kernels. For comparison we also display (black plusses) the 5 observed log-mortality rates during 2014--2019. \emph{Right:} 4 sample paths from 3 representative kernels.
    \label{fig:Age-75-pred}}
\end{figure}

In Figure \ref{fig:Age-75-pred}, we present in-sample and future forecasts of log-mortality for JPN Female Age 65.  The left panel uses the top-10 kernels, providing their posterior mean and prediction intervals. The in-sample fit is tightly constrained, while the out-of-sample prediction becomes more heterogeneous as we move away from the training sample. In particular, there is a bimodal prediction that groups kernels, with some projecting future mortality improvement and others moderating the downward trend. The overlayed 90\% posterior prediction intervals indicate a common region for future mortality trajectories, with symmetric fanning as calendar year increases and a slight skew toward lower mortality rates deeper in future years. This forecasting approach can serve as a basis for Bayesian model averaging, utilizing the Bayes factors as weights. Furthermore, we observe a square-root type fanning of variance, which is common in random-walk mortality models.

The right panel of the figure investigates the stochasticity of the GP by simulating paths using three representative kernels. In-sample paths cluster closely around their posterior means, with observed difficulties in deviating far from the observed data. The observed roughness in the trajectories is a consequence of including a M12 or Min component in Calendar Year or Cohort. When examining the trajectories out-of-sample, the impact of individual kernels becomes more apparent, particularly in the green and blue paths. Lack of mean reversion is more evident in these paths, which could be attributed to the presence of the non-stationary Min kernel in calendar year (green) and Mehler kernel in cohort (blue).

\subsection{Robustness Check}

To validate the above results of the GA, we perform two checks:
(i) re-run the algorithm from scratch, to validate stability across GA runs;
(ii) run the GA on an expanded dataset $\cDrob$, namely we augment with 2 extra calendar years (beginning at 1988 instead of 1990), and 2 extra Age groups (48-86 instead of 50-84).

In both cases, we expect results to be very similar to the ``main'' run discussed above. While the GA undertakes random permutations and has a random initialization, we expect that with 200 kernels per generation and 20 generations, the GA explores sufficiently well that the ultimate best-performing kernels are invariant across GA runs. This is the first justification to accept GA outputs as the ``true'' best fitting kernels. Similarly, while the BIC metric is determined by the precise dataset, we expect that it is sufficiently stable when the dataset undergoes a small modification, so the best fitting kernels can be interpreted as being the right ones for the population in question, and not just for the particular data subset picked.

The above robustness checks are summarized below and in Table \ref{tbl:jpn-robust}. These confirm that the GA is stable both across its own runs (see the ``Re-run" listings) and when subjected to slightly modifying the training dataset. We observe that we recover essentially the same kernels (both in ${\cal K}_r$ and ${\cal K}_f$), and moreover the best kernels/hyperparameters are highly stable as we enlarge the dataset (see the ``Robust'' listings).  This pertains both to the very top kernels, explicitly listed below, but also to the larger set of top-100 kernels, see the summary statistics in Table \ref{tbl:jpn-robust}. Re-assuringly, the kernel length-scales listed below change minimally when run on a larger dataset, confirming the stability of the MLE GP sub-routines. In particular, the same kernel is identified as the best one during the re-run in ${\cal K}_r$, and it shows up yet again as the best for $\cDrob$, with just slightly modified parameters:

\begin{align*}
\text{original $\cD$: } &0.4651 \cdot \M52_a(1.11) \cdot \M52_y(1.80)\cdot \M12_y(62.79) \cdot \M12_c(117.65); \\
\text{enlarged $\cDrob$: } &0.4646\cdot \M52_a(1.11)\cdot \M52_y(1.80)\cdot \M12_y(62.72)\cdot   \M12_c(117.50).
\end{align*}

Four out of the five best kernels repeat when working with ${\cal D}_{rob}$. This stability can be contrasted with \citet{cairns2011mortality} who comment on sensitivity of SVD-based fitting to date range. The other alternatives continue to follow familiar substitution patterns. Of note, with the run over $\cDrob$ there is the appearance of $\Chy_a$, but no appearance of $\Min_y, \M32_y$ or $\Meh_c$. As can be seen in Figure \ref{fig:circBar-jpn}, $\Chy_a$ is actually quite common.

Furthermore, all runs (original, re-run, enlarged dataset) always select $\M52$ or Cauchy kernel for the Age effect, Mat\'ern-1/2 (or sometimes $\Min$) in Period, typically augmented with a smoother kernel like $\M32_y,\Meh_y,\Chy_y$ and  $\M12_c$ in Cohort. 
We furthermore record very similar frequency of different kernels among top-100 proposals, and similar BIC scores for the re-run.

\begin{figure}
   \centering
   \begin{tabular}{cc}
    \includegraphics[width=0.45\textwidth]{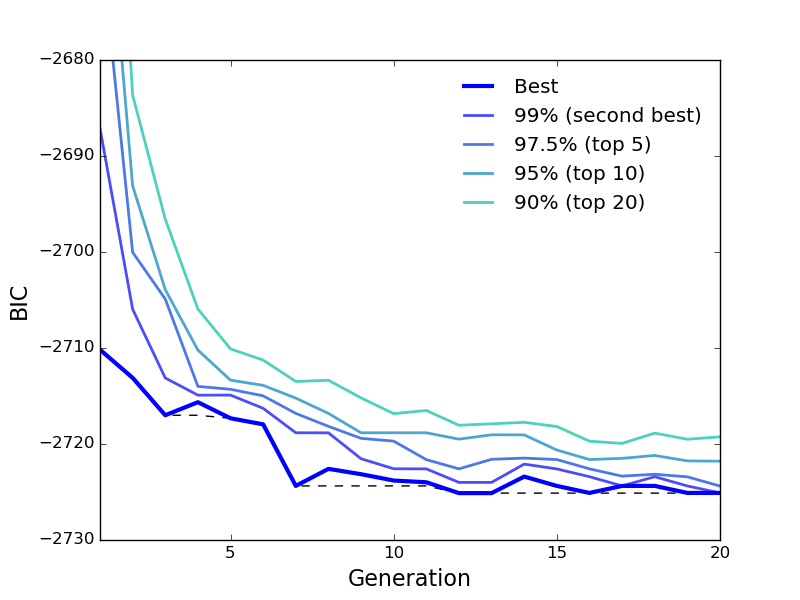} &
    \includegraphics[width=0.45\textwidth]{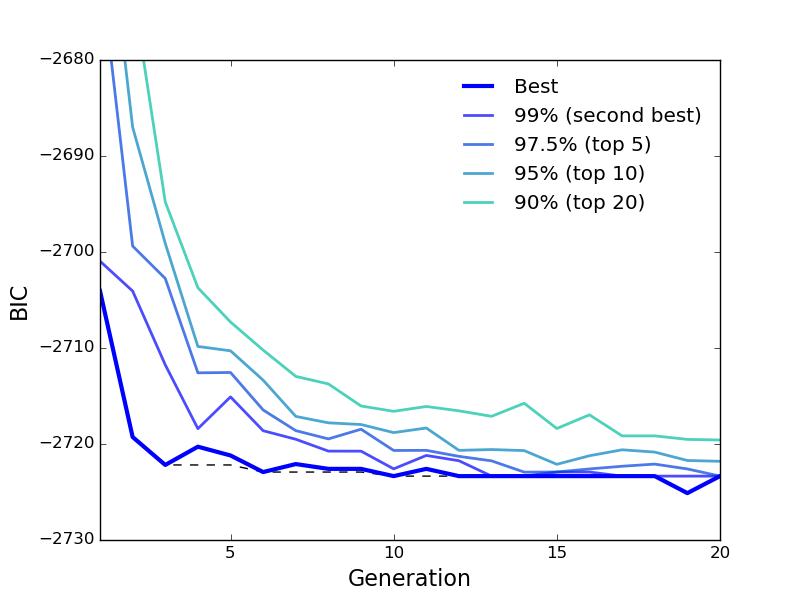} \\
    Main run & Re-run \\
    \end{tabular}
    \caption{Summary statistics of best kernels proposed by GA as a function of generation $g$.
    \label{fig:ga_convergence}}
\end{figure}

As another validation of GA convergence, Figure \ref{fig:ga_convergence} shows the evolution of fitness scores over generations. We display the BIC of the best kernel in generation $g$, as well as the second best (99\% quantile across 200 kernels), 5th best (97.5\%), 10th best (95\%) and 20th best (90\%) across the main run of the GA and a ``re-run''. The experiments in each panel differ only through a different initial seed for the first generation. In both settings, only minimal performance increases (according to the minimum BIC) are found beyond generation 12 or so. Since in each new generation there is inherent randomness in newly proposed kernels, there is only distributional convergence of the BIC scores as new kernels are continuously tried out. This churn is indicated by the flat curves of the respective within-generation BIC quantiles. In sum, the GA converges to its ``equilibrium'' after about a dozen generations, validating our use of $G=20$ for analysis.

\subsection{Analysis of Model Residuals}

\begin{figure}[!ht]
\begin{tabular}{cc}
\includegraphics[width=0.5\textwidth]{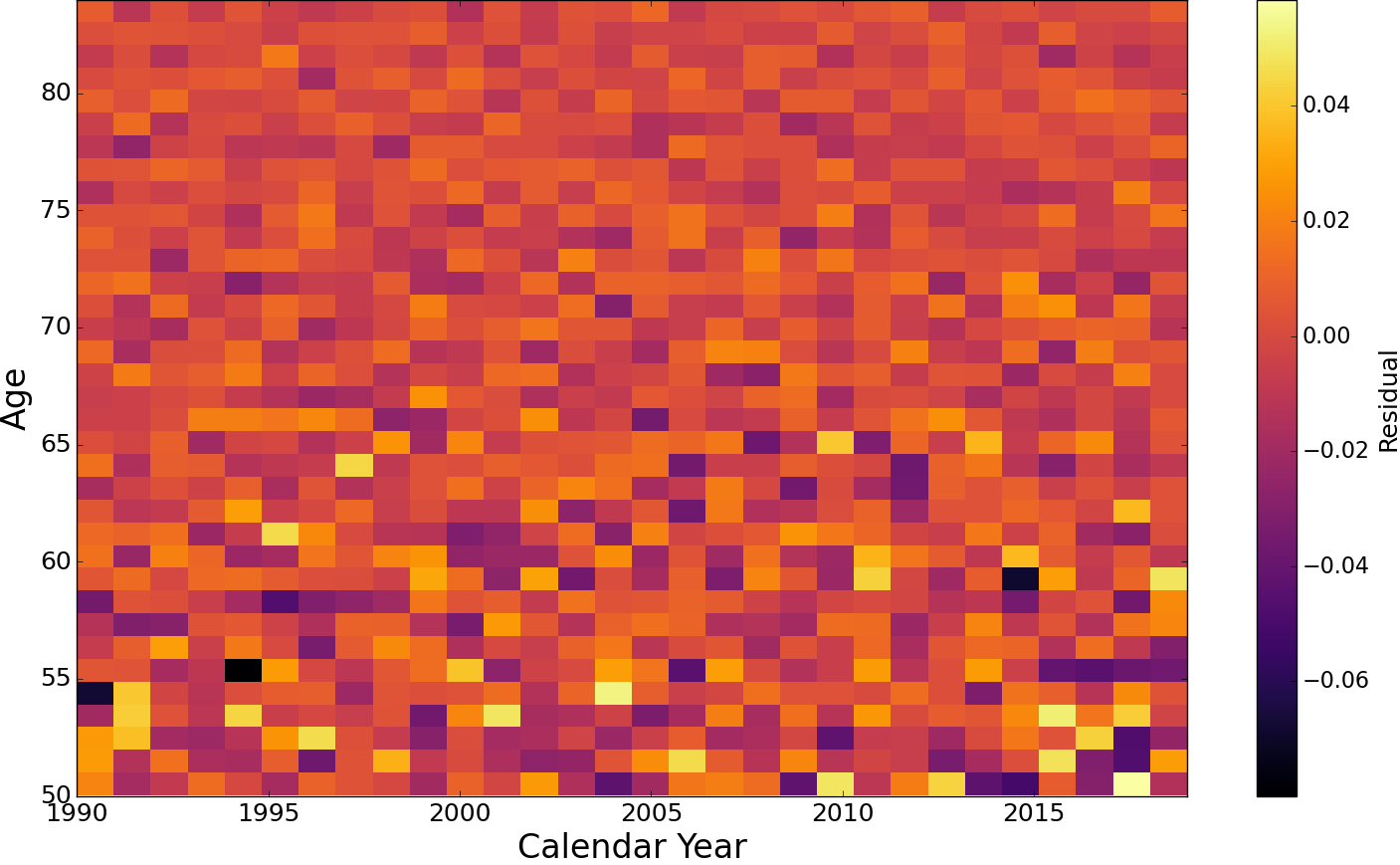} &
\includegraphics[width=0.48\textwidth]{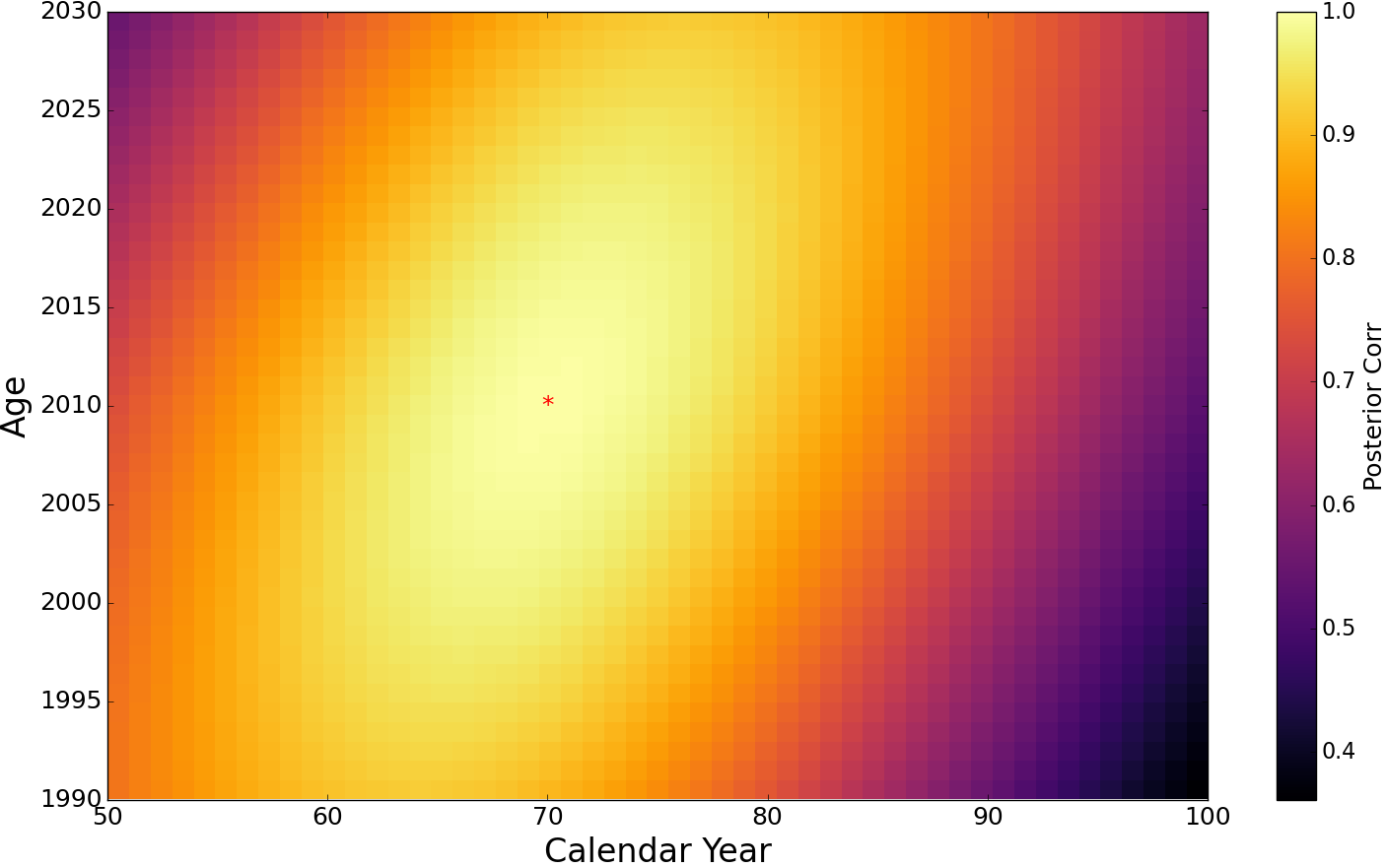}
\end{tabular}
\caption{Right: implied prior correlation of the best kernel. Left: residuals from the best kernel in ${\cal K}_f$ for JPN Females. \label{fig:residuals}}
\end{figure}

In Figure \ref{fig:residuals}, the left panel displays the residuals that compare the realized log-mortality rates of JPN Females with the GP prediction from the best performing kernel. The absence of any identifiable structure, especially along the SW-NE diagonals that correspond to birth cohorts, indicates a statistically sound fit, consistent with the expected uncorrelated and identically distributed residuals. Additionally, we observe distinct heteroskedasticity, where residuals for smaller Ages exhibit higher variance. This is due to the smaller number of deaths at those Ages, resulting in a more uncertain inferred mortality rate, despite the larger number of exposures. Generally, the observation variance is lowest around Age 80.

The right panel of Figure \ref{fig:residuals} shows the prior correlation relative to the cell (70, 2010). The strong diagonal shape indicates the importance of the cohort effect. Moreover, we observe that the correlation decays about the same in Period (vertical) as in Age (horizontal), with the inferred length-scales imposing a dependence of about $\pm 12$ years in each direction.

\subsection{Male vs Female Populations}

We proceed to apply the GA to the JPN Male population. The rationale behind this comparison is the assumption of similar correlation structures between the genders, which enables us to both highlight the similarities and pinpoint the observed differences.

As expected, the JPN Male results (see Table \ref{tbl:results-countries} and Table \ref{tbl:best-kernels}) strongly resemble those of JPN Females. Once again, we detect a strong indication of a single multiplicative term, characterized by APC structure with smooth Age and rough Period and Cohort effects. Just like for Females, a (rough) Cohort term is selected in all (100\%) of the top-performing kernels for JPN Males. The best-fitting individual kernels, as shown in Table \ref{tbl:best-kernels}, are also similar, with M52 in Age, M52 or $\RBF$ in Year, and M12 in cohort being identified as the optimal choices.  

Some differences, such as more Cohort-linked kernels for JPN Males vs.~Females are also observed. The cohort length-scale is much larger for females (120 vs 40), and so is the rough M12\_y length-scale (62  vs 39), while the Age length-scales are almost identical. One interpretation is that there is more idiosyncratic noise in Male mortality, leading to faster correlation decay.

\subsection{Analysis across countries} \label{sec:more-kernel-results}

To offer a broader cross-section of the global mortality experience, we next also consider US males and Sweden females. In total we thus analyze 4 datasets: US males, Japan females and males, Sweden females. We note that Sweden is much smaller (10M population compared to 130M in Japan and 330M in US) than the other 2 countries and therefore has much noisier data.

\begin{table}[ht!]
\begin{tabular}{lrrrrrrrrrrr}
\toprule
Range &  BIC &  BIC &  len &  addtv  &  non- &  num & num & num & rough  &  rough &  rough   \\
  & max & min &  & comps & stat. & age & year & coh & age & year & coh \\
\midrule  \multicolumn{12}{c}{JPN Male} \\ \midrule
 1-10 & -2978.43 & -2980.53 & 4.10 & 1.00 & 0\% & 1.00 & 1.60 & 1.50 & 0\% & 100\% & 100\% \\
 1-50 & -2975.36 & -2980.53 & 4.26 & 1.10 & 0\% & 1.06 & 1.70 & 1.50 & 18\% & 100\% & 100\% \\
51-100 & -2974.25 & -2975.32 & 4.60 & 1.00 & 0\% & 1.04 & 2.14 & 1.42 & 64\% & 100\% & 100\% \\
\midrule  \multicolumn{12}{c}{US Male} \\  \midrule
 1-10 & -3163.54 & -3170.29 & 5.70 & 2.30 & 0\% & 1.50 & 1.50 & 2.70 & 100\% & 100\% & 100\% \\
 1-50 & -3160.32 & -3170.29 & 5.78 & 2.24 & 0\% & 1.40 & 1.54 & 2.84 & 100\% & 100\% & 100\% \\
 51-100 & -3157.93 & -3160.24 & 6.14 & 2.38 & 2\% & 1.46 & 1.72 & 2.96 & 100\% & 100\% & 98\% \\
\midrule  \multicolumn{12}{c}{SWE Female} \\ \midrule
1-10 & -1624.34 & -1625.57 & 3.00 & 1.00 & 0\% & 1.00 & 1.00 & 1.00 & 0\% & 100\% & 0\% \\
 1-50 & -1622.74 & -1625.57 & 3.02 & 1.00 & 6\% & 1.00 & 1.24 & 0.78 & 0\% & 100\% & 14\% \\
 51-100 & -1622.04 & -1622.74 & 3.42 & 1.04 & 16\% & 1.10 & 1.38 & 0.94 & 0\% & 100\% & 6\% \\
\bottomrule
\end{tabular}
\caption{Results from GA runs on JPN Male, US Male and SWE Female. Throughout we search within the full set ${\cal K}_f$. \label{tbl:results-countries}}
\end{table}

\begin{table}
\begin{tabular}{r|r p{4.5in}} \toprule
Pop'n/Search Set & $N_{pl}$ & Top Kernel \\ \midrule
JPN Female ${\cal K}_r$ & 90  &  $0.464\cdot \M52_a(1.1) \cdot \RBF_y(1.33) \M12_y(62.51) \cdot \M12_c(118.06) $ \\
JPN Female ${\cal K}_f$ & 95 & $0.4638 \cdot \M52_a(1.11) \cdot \Chy_y(1.95) \M12_y(62.42)\cdot  \M12_c(117.11)$ \\
JPN Male ${\cal K}_r$ & 89 & $0.1491\cdot \M52_a(0.95)\cdot \RBF_y(1.15) \M12_y(26.24)\cdot \M12_c(24.90)$ \\
JPN Male ${\cal K}_f$ & 112 & $0.2130\cdot \M52_a(1.09)\cdot \M12_y(39.09)\cdot M32_c(0.86) \M12_c(40.73)$ \\
US Male ${\cal K}_r$ & 57 & $0.017\cdot \M12_a(5.04) \cdot  \M52_y(0.50) M12_y(10.33)\cdot \M52_c(0.36) \M12_c(5.00)$ \\
US Male ${\cal K}_f$ & 35 &  $0.01 \cdot \AR2_a(1.12,1.88) \cdot \M12_y(24.18)  \cdot \M32_c(0.72)\cdot [4.6211\cdot \M12_c(13.49) + 0.01  \cdot \M32_a(0.02) \cdot \M52_c(0.1)]$ \\
SWE Female ${\cal K}_r$ & 200+ & $0.2527\cdot \RBF_a(0.52)\cdot \M12_y(73.74)\cdot \RBF_c(0.62)$ \\
SWE Female ${\cal K}_f$ & 200+  &$0.2094\cdot \Chy_a(1.05) \cdot \M12_y(67.27)\cdot \Meh_c(0.60) $ \\ \bottomrule
\end{tabular}
\caption{\label{tbl:best-kernels} Best performing kernel in ${\cal K}_r$ and ${\cal K}_f$ for each of the 4 populations considered. $N_{pl}$ is the number of alternate kernels that have a BIC within 6.802 of the top kernel and hence are judged ``plausible'' based on the BF criterion. }
\end{table}

\textbf{USA Males:} The US male data  leads to kernels of much higher length compared to all other datasets. The GA returns kernels with 5-6 base kernels, and frequently includes two or even three additive terms. Moreover, the APC pattern is somewhat disrupted, possibly due to collinearity between the multiple Period and Cohort terms.

To demonstrate some of the observed characteristics, let us examine the top kernel in ${\cal K}_f$, as presented in Table \ref{tbl:best-kernels}. This kernel comprises 11 terms, including 2 additive terms. However, we note that the second term has a significantly smaller coefficient, indicating that it serves as a "correction" term that is likely introduced to account for a less prominent and identifiable feature generally overshadowed by the primary term. Additionally, we observe that this kernel incorporates both rough Cohort term $\M12_c$ and smoother ones, namely $\M32_c$ and $\M52_c$. This points towards a multi-scale Cohort effect, where a few exceptional years (such as birth years during the Spanish Flu outbreak in 1918-1919) are combined with generational patterns (e.g., Baby Boomers vs. the Silent Generation). Unlike other datasets, the US Male data even includes a $\RBF_c$ term. Finally, the Age effect is described by the $AR2$ kernel, which is also commonly observed in the JPN Male population.

Moving down the list, there are also shorter kernels with a single additive component, for example $0.0129 \cdot \M12_a(3.44) \cdot (\RBF_y(0.63) \M12_y(7.89)) \cdot (\M32_c(0.43)  \M12_c(3.88)$ which is fourth-best and the length-7 $0.0113 \cdot \M12_a(2.61) \cdot (\M32_y(0.7)  \M12_y(6.53))\cdot \M12_c(2.91)$ which is seventh-best.
In all, for US males we can find a plausible kernel of length 7, 9, 11, 13 when the best performing one has length 11.
This wide distribution of plausible kernel lengths (and a wide range of proposed kernel families) is illustrated in the right panel of Figure \ref{fig:kernel-length} and the middle column of Figure \ref{fig:circBar}. 

Another sign that the US data has inherent complexity is the wide gap in BF of the best kernel in ${\cal K}_f$ compared to that in ${\cal K}_r$, by far the biggest among all populations. Thus, restricting to ${\cal K}_r$ materially worsens the fit. In fact, we observe that all the top kernels in ${\cal K}_r$ are purely multiplicative (such as $\M12_a  \cdot \M52_y \M12_y \cdot \M52_c \M12_c$), which is unlikely to be the correct structure for this data and moreover hints at difficulty in capturing the correlation in each coordinate, leading to multiple Period and Cohort terms. Within ${\cal K}_f$ only 35 plausible alternatives are found, 3-5 times fewer than in other datasets.

\bigskip

\textbf{SWE Female:}
The Swedish Females dataset turns out to have two distinguishing features. First, it yields the simplest and shortest kernels that directly match the APC structure of 3 multiplicative terms. The average kernel length reported in Table \ref{tbl:results-countries} is the smallest for SWE Females, and additive terms appear very rarely. When kernels with more than 3 terms are proposed, these are usually still all-multiplicative, and add either a second Period term (e.g  $\Meh_a \cdot \M12_y \cdot \Meh_y \cdot  \Meh_c$) or a second Age term, though both are smooth: $(\Chy_a \RBF_a) \cdot  \M12_y \cdot \Meh_c$).

Second, and unlike all other populations above, the Cohort effect is ambiguous in Sweden. About 15\% of the top performing kernels (and 30\% in ${\cal K}_r$) have no Cohort terms at all, instead proposing two terms for the Period effect. Those that do include a Cohort effect, use either a $\RBF_c$ or a $\Meh_c$ term, indicating no short-term cohort features, but only generational ones. Nine of the top-10 kernels in ${\cal K}_f$ and only 7 out of 10 in ${\cal K}_r$ have Cohort terms. In contrast, in JPN and US rough cohort terms are present in every single top-100 kernel. Once again, our results are consistent with the literature; see \citet{murphy2010reexamining} who discusses the lack of clarity on cohort effect for the Swedish female population.

A third observation is that SWE Female shows a compression of BIC values, i.e.~a lot of different kernels are proposed with very similar BICs.  The GA returns over two hundred kernels with a BF within a ratio of 30 to the top one. This indicates little evidence to distinguish many different choices from each other and could be driven by the lower-complexity of the Swedish mortality data.

\subsection{Discussion}

\begin{figure}[ht]
\hspace*{-20pt} \vspace*{-4pt}
\begin{tabular}{ccc}
 \includegraphics[height=2.2in,trim=0.5in 0.1in 0.4in 0.1in]{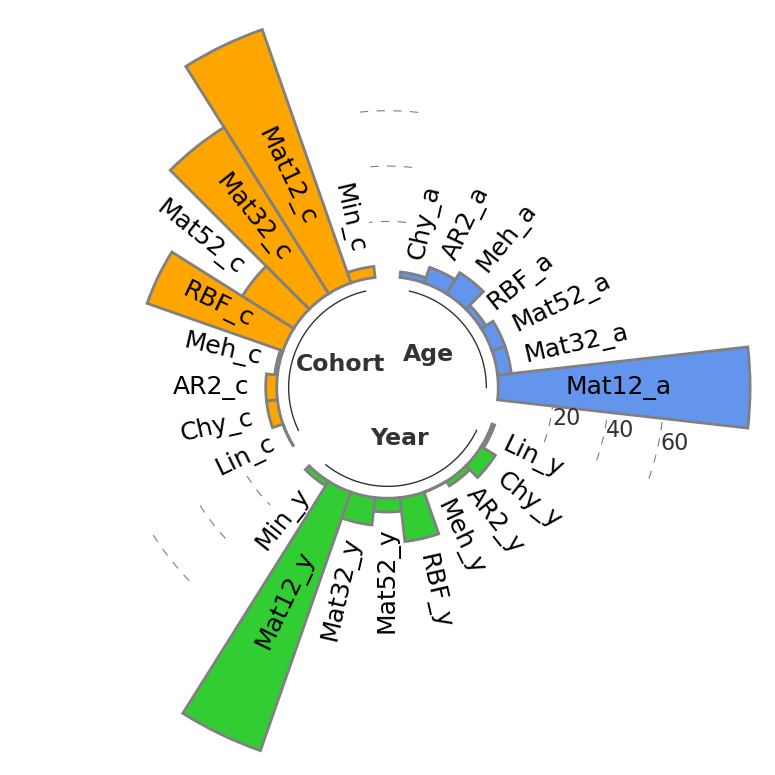} &
\includegraphics[height=2.2in,trim=0in 0.4in 0.4in 0.1in]{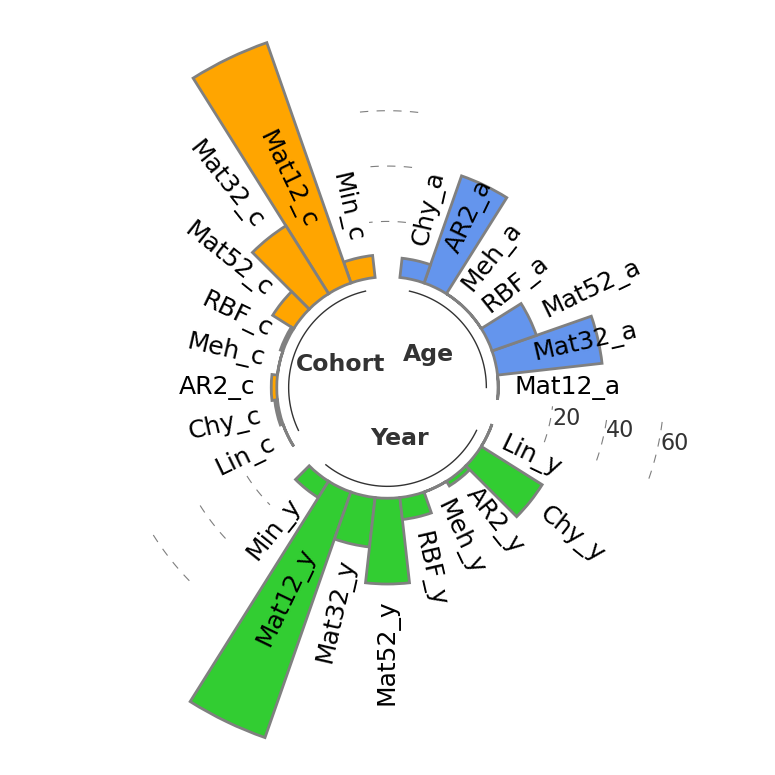} &
\includegraphics[height=2.2in,trim=0.3in 0.7in 0.7in 0.1in]{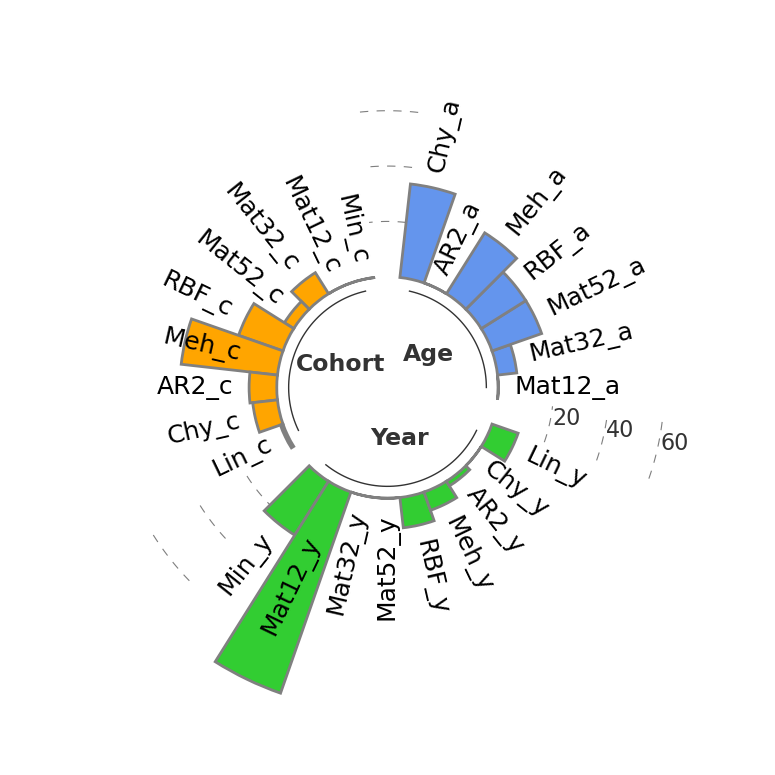} \\
US Male  & JPN Male  & SWE Female
\end{tabular}
\caption{Frequency of appearance of different kernels from ${\cal K}_f$ in US, SWE and JPN Male models. \label{fig:circBar}}
\end{figure}

\textbf{Best kernel families:}
The barplots in Figure \ref{fig:circBar} suggest that there is no clear-cut covariance structure that fits mortality patterns. Consequently, the models often propose a combination (usually a product, sometimes a sum) of various kernels. Moreover, there is no one-size-fits-all solution as far as different populations are concerned. For instance, the Age effect is typically modeled via a $\M12$ kernel for US Males, a $\M32$ kernel for JPN Males, and a $\Chy$ (or RBF or $\M52$) kernel for SWE Females. Additionally, $\Chy$ may be identified as a possible Period term for US and JPN Males but never for SWE Females. As such, it is recommended to select different kernels for different case studies. This represents one of the significant differences compared to the classical APC framework, where the SVD decomposition is invariant across datasets, and researchers must manually test numerous combinations, as illustrated in \citet{cairns2011mortality}.

\textbf{Necessity of Cohort Effect:}

To assess the impact of including a Birth Cohort term, we re-run the GA while excluding all cohort-specific kernels. This is a straightforward adjustment to the implementation and can be used to test whether cohort effects could be adequately explained through a well chosen Age-Period kernel combination.  

We first evaluate our models by comparing the Bayesian Information Criterion (BIC) of the top kernel that excludes Cohort terms with that of the full ${\cal K}_f$. Additionally, we examine the residuals heatmap to detect any discernible diagonal patterns. Our findings indicate that the Cohort effect is overwhelmingly needed for US Males, JPN Females, and JPN Males. The absence of a Cohort term results in a significant increase in BIC, with a difference of 235 for JPN Male, 198 for US Male, and 111 for JPN Female. To put things in perspective, a BIC difference is considered significant only when it surpasses 6.802. The results are confirmed by Figure \ref{fig:resid_nocohort}, which illustrates pronounced diagonals in the bottom row (the no-cohort models), contradicting the assumption of independent residuals.

For Sweden, the difference is only 1.24, which corresponds to a BF of 0.2894, indicating no significance. Moreover, the corresponding no-cohort residuals still appear satisfactory, and the associated kernel $0.1125\cdot \Chy_a(0.85) \cdot \Meh_y(1.16)\cdot \M12_y(37.74)$ is ranked ninth-best in the original ${\cal K}_f$. Once again, our results are consistent with the literature; see \cite{murphy2010reexamining} who discusses the lack of clarity on cohort effect for Sweden.

\citet{cairns2011mortality}  suggested that cohort effects might be partially or completely explained by well-chosen age and period effects. We find a partial confirmation of this finding in that in many populations there are more than 2 kernels used to explain the Period and Cohort dependence, and there is a clear substitution between them. However, this is a second order effect; the primary necessity of including Cohort is unambiguous except for Sweden. For Sweden females, the need for birth cohort dependence is quite weak.

\textbf{Kernel Substitution:}

Through its mutation operations, the GA naturally highlights substitution effects among different kernel families. Substitution of one kernel with another is intrinsic to the evolution of the GA, and by ranking the kernels in terms of their BIC, we observe the presence of many compositions that achieve nearly same performance and differ just by 1 term. The above effect is especially noticeable in purely multiplicative kernels that are prevalent in all populations except the US. In this case, we can frequently observe that one of the terms can be represented by two or three kernel families, with the rest of the terms staying fixed.

We observed that certain kernels are commonly used as substitutes for each other, such as the $\RBF$ and $\M52$ kernels, as well as the $\M12$ and $\Min$ kernels. Although the latter pair differs by stationarity, the sample paths generated by $\Min$ are visually indistinguishable from those generated by $\M12$ when the length-scale parameter $\ell$ is large (usually $\ell_{\Min} > 25$). These substitution patterns are in agreement with our synthetic results, as discussed in Section \ref{sec:synthetic}.  As an example, when training on the Japan Female dataset and searching in ${\cal K}_r$, the top kernel is of the form $\M52_a \cdot (\RBF_y \cdot \M12_y) \cdot \M12_c$, while the second-best according to BIC is $\M52_a \cdot (\M52_y \cdot M12_y)\cdot M12_c$. This preserves the same macro-structure while replacing one of the two Period terms; cf. Table \ref{tab:JPN-best-results}.
Additionally, the next two ranked kernels are very similar, but substitute $\M12$ with $\Min$: the third-best is $\M52_a \cdot (\RBF_y \cdot \Min_y) \cdot \M12_c$, and the fourth-best is $\M52_a \cdot (\M52_y \cdot \Min_y) \cdot \M12_c$. It is important to note that during substitution, the lengthscales (and sometimes process variances) change, as these have a different meaning for different kernel families. For example, the $\RBF$ lengthscales tend to be about 50\% smaller than those for its substitute, $\M52$.

The Cauchy kernel and $\M52$ are also interchangeable, although $\Chy$ and $\RBF$ are less so. This effect of multiple substitutes for smooth kernels is nicely illustrated for SWE Females, where the four top kernels fix the Period and Cohort effects according to $\M12_y \cdot \Meh_c$ and then propose any of $\Chy_a, \RBF_a, \Meh_a, \M52_a$ for the Age effect. Substitutions are more common within ${\cal K}_f$, since the availability of more kernel families contributes to ``collinearity''  and hence more opportunities for substitution.
At the same time, we occasionally observe the ability to find a more suitable kernel family in ${\cal K}_f$. For example often we observe both a rough and a smooth kernel in Year,  indicating that neither M12 nor RBF fit well on their own; in that case $\M32$ sometimes appears to be a better single substitute. Similarly, $\Meh_c$ is the most common choice for SWE Females, and is replaced with $\RBF_c$ or $\RBF_c \cdot \M12_c$ in ${\cal K}_r$. Consequently, proposed kernels from ${\cal K}_f$ tend to be a bit shorter on average than those from ${\cal K}_r$.

A further substitution effect happens between Period and Cohort terms. In JPN Females and SWE Females, we tend to observe a total of three terms, and there is a substitution between using two Period and one Cohort or one Period and two Cohort terms.
For instance, in JPN female among top-10 kernels we find both
$\M52_a \cdot \M12_y \cdot (\Meh_c \cdot \M12_c)$ and
$\M52_a \cdot \Meh_y \M12y \cdot \M12_c$.

\textbf{Additivity and Nonstationarity:}

Returning to the topic of additive components, our results generally suggest that the additive structure is  is generally weak. Specifically, introducing an additional additive component often provides only a minor improvement in goodness of fit, which is offset by the complexity penalty in BIC, resulting in lower BICs for additive kernels. Hence, additive kernels tend to be rejected by the GA on the grounds of parsimony.  As a result, most kernels with 4 terms (length 7) and the majority with 5 terms are purely multiplicative.

Summarizing non-stationarity is a challenging task due to various factors, so any table providing the percentage of non-stationary data should be viewed with reservation. A more comprehensive analysis of non-stationarity is can be seen through the frequency diagrams shown in Figures \ref{fig:circBar-jpn} and \ref{fig:circBar}. It is important to note that M12 lengthscales tend to be large in fitting. When M12 has a large lengthscale, the resulting processes are visually indistinguishable from those generated by the non-stationary Min kernel. Therefore, in our data, the presence of M12 indicates potential non-stationarity. From this perspective, our findings suggest that all populations exhibit a (potentially) non-stationary period effect. The other non-stationary kernels (Lin, Meh) are rare but do occur, mostly in the SWE female population.

\section{Conclusion} \label{sec:conclude}

Our work analyzes the use of a genetic algorithm to discover kernels (i.e.~covariance
structures) for Gaussian process surrogates of mortality surfaces.  The GA performed excellently in our synthetic experiments, indicating its promising role as a tool for model selection and validation when using the GP framework for realistic data analysis.  In particular, it successfully detected the smoothness of the data generating process, demonstrated robustness across samples (SYA), distinguished additive versus multiplicative APC structures, identified relatively small cohort effects (SYB), and found the correct number of base kernels and identified multiple nonstationarities over Period and Cohort (SYC) coordinates. Additionally, all experiments illustrate the ``substitution" effect, where one kernel approximates the impact of another. For instance, $\M12$ kernel with large $\ell_{\text{len}}$ can substitute for $\Min$ and vice versa.

When applied to the HMD datasets, our results strongly suggest that best fits to mortality data are provided by GP models that include a rough (non-differentiable or only once-differentiable) component in Year and Cohort, and smooth terms in Age. This matches the classical assumption that the Age-structure of mortality is a smooth function, while the temporal dynamics are random-walk-like. The only exception is the US data, where Age structure is proposed to be non-differentiable, while the Cohort term is smooth.

Among non-standard kernels, we find that Cauchy kernels are often picked, with $\Chy_a$ showing up in 34\% of SWE Female and 28\% of JPN Female top-100, and $\Chy_y$ in 26\% of JPN Male. Mehler kernels also appear, though infrequently except for $\Meh_c$ in SWE females (35 out of top 100 kernels).

Historical data analysis and the SYC experiment both revealed a lack of clarity on determining one single covariance structure. This is unsurprising, given the presence of surrogate kernels that mimic one another (e.g.~$\Chy_a$ instead of $\M52_a$, or $\Meh_c$ instead of $\RBF_c$) and the complexity of the search problem when using the larger search set $\cK_f$ (twice as many kernels). Although this benefits BIC optimization by better approximating the truth, finding a precise and expressive covariance structure requires a smaller  $\cK$ that includes key families that express the modeler's prior beliefs about the underlying data process. This is a challenging knowledge to have as it requires expertise in the application area (e.g.~mortality modeling) and properties of Gaussian process kernels.

One of our foci in this article was examining the kernels of the best-fitting models. In parallel, the GA output also supports model averaging analysis, i.e., the simultaneous use of multiple models. Model averaging provides insight into model risk and robustifies predictions against the idiosyncrasies of any given kernel choice. It thus offers a high-powered alternative to Bayesian GPs, as applied in \cite{huynh2021multi}.  Due to space limitations, this is left to a separate project.

Additional avenues for future work include transferring our approach to smaller populations (under 10-20 million).  In contrast to our data where the signal to noise ratio was high and hence the GA had a favorable setting to infer the best latent structure, this is likely to require a significant adjustment since with higher noise the GA will have a harder time differentiating candidate kernels. One solution would be to build a version that jointly models multiple populations using multi-output Gaussian Processes \cite{huynh2021joint,huynh2021multi}.

Additionally, further kernels can be considered.  One example is that the Cauchy kernel is a special case of the rational quadratic kernel indexed by $\alpha$ (see Appendix \ref{sec:kernel_appendix}).  Given the popularity of $\Chy$ in our results, it may be worthwhile to explore $\alpha$ values beyond $\alpha=1$, or consider a direct search over $\alpha$.  It is also possible to incorporate a Cohort effect into non-separable kernels over Age and Year dimensions, where $x_{yr}-x_{ag}$ can naturally appear in expressions such as $\exp(-[x_{ag} , x_{yr}]^\top A [ x_{ag} , x_{yr} ])$ to define a kernel when $A$ is positive definite.

Changepoint detection can be naturally implemented with GPs, utilizing as kernel $\sigma(x) k_1(x,x')\sigma(x') + \overline{\sigma}(x) k_2(x,x') \overline{\sigma}(x')$, where $\overline{\sigma}(x) = 1-\sigma(x)$ and $\sigma(\cdot)$ is an activation function, like the sigmoid $\sigma(x) = 1/(1+\exp(-x))$.  This is useful when there is a non-stationary shift from one mortality structure to another, i.e.~the transitioning from younger to older ages, or in the presence of a temporal mortality shift. Rather than prescribing a hard cutoff, one could design kernels that automatically explore that possibility.

Furthermore, carefully chosen hyperparameter constraints could aid in finding better structures.  In the additive case, an additional term always penalizes BIC through an additional scale coefficient.  It may make sense to add a term that uses the same scale coefficient (i.e.~factoring), or a reduction of an existing scale factor by a fixed amount (e.g.~0.5 or 0.25).  Through re-using existing hyperparameters, this would not penalize BIC.  There may also be ways to constrain relations between Period and Cohort lengthscales for better interpretability.

Lastly, more work is warranted to understand the limitations of the GA as currently built. Additional analysis of synthetic experiments can help clarify the impact of signal-to-noise ratio and the size of the dataset on the ability of the GA to appropriately explore in order to find the best-fitting kernel. Similarly, we leave it to a future study the analysis of whether it is always better to ``throw in the kitchen sink'', as far as including as many diverse kernels as feasible, or whether a pre-selection could be beneficial.

\bibliography{biblio}

\appendix

\section{Notes on Kernels}\label{sec:kernel_appendix}

Unless otherwise stated, the following assumes $x, x' \in \R$.  
In this section we use the fact that for a GP $f$, its derivative $f'$ exists (in the mean-square sense) if and only if $\frac{\partial^2 k}{\partial x \partial x'}(x, x')$ exists \cite{adler2010geometry}.

\subsubsection*{Stationary}

\begin{itemize}
\item[Mat\'ern-1/2:] is the covariance of an Ornstein-Uhlenbeck (OU) process \cite{berlinet2011reproducing}. The OU process follows a linear mean-reverting stochastic differential equation; it has continuous, nowhere differentiable paths. The mean-reversion localizes dependence and has been advocated \cite{jahnichen2018scalable} for capturing small-scale effects. The lengthscale $\ell_\text{len}$ controls the rate of mean-reversion (lower values revert more quickly).

The re-parametrization
\begin{equation}
    k(x,x') = \exp\left(-\frac{-|x-x'|}{\ell_{\text{len}}}\right) = \phi^{|x-x'|}, \qquad \phi = \exp(-1/\ell_\text{len}),
\end{equation}
shows that when $x$ is discrete, $\M12$ is equivalent to an AR(1) process with persistence parameter $\phi$. In particular,  $\ell_\text{len}$ large (i.e.~$\phi \simeq 1$)
mimics the non-stationary random walk process and its $\Min$ kernel, allowing sample paths to deviate far from their mean and weakening stationarity.

 \item[AR2:]
 The covariance kernel associated with a (continuous $x$) second order autoregressive (AR(2)) process is
 \cite{parzen1961approach}
 \begin{equation*}
     k(x,x'; \alpha, \gamma) = \frac{\exp(-\alpha |x-x'|)}{4\alpha \gamma^2}\left\{\cos(\omega |x-x'|) + \frac{\alpha}{\omega} \sin(\omega |x-x'|)\right\},
 \end{equation*}
 where $\omega^2 = \gamma^2 - \alpha^2 > 0$. Notably, this kernel has two parameters and, thus, a higher BIC penalty during the GA optimization.  One can show that $\frac{\partial^2 k}{\partial x \partial x'}(x, x')$ exists for all $x, x' \in \R$, but $\frac{\partial^4 k}{\partial^2 x \partial^2 x'}(x, x')$ does not.  Thus a GP with AR2 kernel is (mean-square) once but not twice differentiable, i.e.~$f \in C^1$.

 For consistency with the Mat\'ern family of kernels, we reparameterize according to $\alpha = 1/\ell_{\text{len}}, \omega = \pi/p$ and normalize for $k(x,x;\ell_{\text{len}}, p)=1$, so that

  \begin{equation}
     k(x,x'; \ell_{\text{len}}, p) = \exp\left(-\frac{|x-x'|}{\ell_{\text{len}}}\right) \big\{\cos\big(\frac{\pi}{p} |x-x'|\big) + \frac{p}{\pi \ell_{\text{len}}} \sin\big(\frac{\pi}{p} |x-x'|\big)\big\},
 \end{equation}

 where, under the re-parametrization, $\gamma^2 = \omega^2 + \alpha^2 = \frac{1}{\ell_{\text{len}}^2} + \frac{\pi^2}{p^2}.$  Through trigonometric identities, one can see that this is the same covariance function as a stationary discrete-time AR(2) process (written as $f(x) = \phi_1 f(x-1) + \phi_2 f(x-2) + \epsilon(x)$) in the case of complex characteristic roots, i.e.~$\phi_1^2 + 4\phi_2 < 0$, with parameters related by
 \begin{equation}
    \phi_2 = -\exp(-2/\ell_{\text{len}}), \qquad
     \phi_1 = 2\cos(\pi/p)\sqrt{-\phi_2}.
 \end{equation}

 \item[Cauchy:] This kernel is fat-tailed and has long-range memory, which means that correlations decay not exponentially but polynomially, leading to a long range influence between inputs \cite{jahnichen2018scalable}. The $\Chy$ kernel function is given by
 \begin{equation}\label{eq:cauchy}
     k(x,x'; \ell_{\text{len}}) = \frac{1}{1+ \frac{(x-x')^2}{\ell_{\text{len}}^2}}
 \end{equation}
 which is a special case, $\alpha=1$, of the rational quadratic kernel $k(x, x'; \alpha, \ell_{\text{len}}) := \left(1+\frac{(x-x')^2}{\alpha\ell_{\text{len}}^2}\right)^{-\alpha}$.

 Since $k(x,x')$ in \eqref{eq:cauchy} is infinitely differentiable in both arguments, the associated GP is also in $C^\infty$ like $\RBF$.  One way to interpret $\Chy$ is as a marginalized version of the RBF kernel with an exponential prior on $1/\ell_{\RBF}^2$ (with rate $\ell_{\Chy}^2/2$):
 \begin{equation*}
     \int_0^\infty \exp\left(-u \cdot \frac{(x-x')^2}{2}\right) \cdot \ell_{\Chy}^2 \exp(-\ell_{\Chy}^2 \cdot u)du = \frac{1}{1+ \frac{(x-x')^2}{\ell_{\Chy}^2}}
 \end{equation*}

\end{itemize}

\subsubsection*{Nonstationary}

\begin{itemize}

\item[Linear:] the linear kernel connects Gaussian processes to Bayesian linear regression.  In particular, if $f(x) = \beta_0 + \beta_1 x$ where $x \in \R$ and there are priors $\beta_0 \sim N(0, \sigma_0^2)$, $\beta_1 \sim N(0, 1)$, then $f \sim \GP(0, k_{\Lin})$, that is, $k(x,x') = \sigma_0^2 + x \cdot x'$.  Note that $k_{\mathrm{Lin}}$ can be scaled to yield a prior variance on $\beta_1$.

 \item[Mehler:]
The nonstationary
 \citet{mehler1866ueber} kernel is

\begin{align}\notag
k(x,x'; \rho) & =  \exp\left(-\frac{\rho^2(x^2 + x'^2) - 2 \rho xx'}{2(1-\rho^2)}\right), \qquad -1 \leq \rho \leq 1 \\
  & = k_{RBF}(x,x'; \ell_{\text{len}}) \cdot \exp\left(\frac{\rho}{\rho+1}xx'\right)
 \end{align}

where $\ell_{\text{len}}^2 = \frac{1-\rho^2}{\rho^2}$. Thus we can interpret the Mehler kernel as another $C^\infty$ kernel that provides a nonstationary scaling to RBF.  Initial experiments always found $\rho > 0$, which causes an increase in covariance for larger values of $x$ and $x'$.  One way to see the effect of the nonstationary component is through variance and correlation.  If $f \sim \GP(m, k_{\Meh})$, then $\var(f(x)) = k_{\Meh}(x,x) = \exp\left(\frac{\rho}{\rho+1}\cdot x^2\right)$ which illustrates an increase in process variance as $x$ increases.  Remarkably, this results in a stationary \emph{correlation function} $\text{corr}(f(x), f(x')) = \exp\left(-\frac{\rho}{2(1-\rho^2)}(x-y)^2\right)$.  Thus, Mehler is appropriate when one desires RBF dynamics combined with increasing process variance.

\begin{rem}
The Mehler kernel is a valid kernel function for $\rho>0$ in the sense that it is positive definite, as it admits the basis expansion $k(x,x') = \frac{1}{\sqrt{1-\rho^2}}\sum_{k=0}^\infty \frac{\rho^k}{k!} H_{e_k}(x) H_{e_k}(x')$ and hence for all $n \in \N$, $\mathbf{a} \in \R^n$ and $x_1, \ldots, x_n \in \R$,
\begin{equation*}
\sum_{i=1}^n \sum_{j=1}^n a_i a_j k(x,x') = \frac{1}{\sqrt{1-\rho^2}} \sum_{k=0}^\infty \rho^k \left(\sum_{i=1}^n a_i h_k(x_i)\right)^2 \geq 0,
\end{equation*}
where $H_{e_k} = (-1)^k e^{x^2/2} \frac{d^k}{dx^k}e^{x^2/2}$ is the $k$th probabilist's Hermite polynomial.
\end{rem}

\end{itemize}

\section{More on GP Hyperparameter Convergence}\label{sec:appendix_convergence}

The mutation operations of our Genetic Algorithm depend crucially on the relative comparison of the log-likelihoods of the given dataset across different kernels $k$'s given its direct role in computing BIC. Hence, accurate estimate of $l(k)$ is a pre-requisite to identify which kernels are fitter than others and hence explore accordingly. Computing the likelihood is equivalent to inferring the MLE for the kernel hyperparameters and is known to be a challenging optimization task. In our implementation, this optimization is done via stochastic gradient descent (SGD) through Adam \cite{kingma2014adam}, up to a given number $\eta_{max}$ of iterations or until a pre-set tolerance threshold is reached.

In this section we present additional evidence on how fast this convergence occurs in our synthetic experiments. Namely, we evaluate GP hyperparameter convergence during training by fitting each of the true kernels $K_0$ from initialization for SYA, SYB, and SYC, indexing  the intermediate log marginal likelihoods after $\eta$ steps as $l^{(\eta)}_{k}(\hat{\theta}| \mathbf{y})$. Given a gold-standard $l_{\text{min}} = \min_{1 \leq \eta} l^{(\eta)}_{k}(\hat{\theta}| \mathbf{y})$,
we record the number of training steps needed to achieve a log marginal likelihood within $\varepsilon \in \{10^{-3}, 10^{-4}, \ldots, 10^{-7}\}$ of $l_{\text{min}}$:
\begin{equation}
    \eta_{\varepsilon} = \min\left\{\eta : |l^{(\eta)}_{k}(\hat{\theta}| \mathbf{y}) - l_{\text{min}}| \leq \varepsilon\right\}.
\end{equation}

We present the results in Table \ref{tab:conv_tol}. SYA-1 and -2 show inconsistency for $\varepsilon=10^{-7}$ probably because the SGD optimization hyperparameters were calibrated to the $\varepsilon = 10^{-6}$ case. SYB takes four times as many training steps as SYA for $\varepsilon = 10^{-4}$, and ten times as many for $\varepsilon = 10^{-6}$. Learning the hyperparameters of SYC is  approximately ten times slower than for SYB for both $\varepsilon = 10^{-4}$ and $10^{-6}$. 
In sum, the convergence to $l_{\text{min}}$ is rapid for SYA and SYB, but very slow for SYC, in part due to the latter having more hyperparameters, non-stationary terms, and non-constant variance.

For computational tractability, in our experiments we have restricted to $\eta_{max} = 300$ (running time is linear in $\eta_{max}$), with the result that we achieve a tolerance of better than $10^{-5}$ for SYA, better than $10^{-4}$ for SYB, but can mis-estimate $l_k$ up to $\pm 1$ for SYC. This provides a rationale for the difficulties encountered in recovering $K_0$ during the SYC experiment, since the GA has trouble correctly sorting the top kernels.

\begin{table}[]
    \centering
\begin{tabular}{c|c|rrrrr}
   & $\varepsilon$ & $10^{-3}$ &  $10^{-4}$ &  $10^{-5}$ &  $10^{-6}$ &  $10^{-7}$ \\
\toprule
  \multirow{ 2}{*}{SYA-1} & $\eta_{\varepsilon}$ & 6 &  62 &  89 &  90 &  90   \\
   & $\BIC(\hat{K}_0^{(\eta_{\varepsilon})})$ & -2066.14 & -2066.67 & -2066.76 & -2066.77 & -2066.77   \\ \midrule
  \multirow{ 2}{*}{SYA-2} & $\eta_{\varepsilon}$ &    39 &    57 &   143 &   179 &  3241   \\
   & $\BIC(\hat{K}_0^{(\eta_{\varepsilon})})$ & -2033.23 & -2034.17 & -2034.22 & -2034.23 & -2034.23   \\ \midrule
  \multirow{ 2}{*}{SYB} & $\eta_{\varepsilon}$ &   165.00 &   239.00 &   603.00 &   835.00 &   997.00  \\
   & $\BIC(\hat{K}_0^{(\eta_{\varepsilon})})$ & -2467.19 & -2467.96 & -2468.06 & -2468.07 & -2468.07    \\ \midrule
  \multirow{ 2}{*}{SYC} & $\eta_{\varepsilon}$ &   669.00 &  2356.00 &  4500.00 &  6553.00 &  8318.00 \\
   & $\BIC(\hat{K}_0^{(\eta_{\varepsilon})})$ & -2794.16 & -2795.11 & -2795.20 & -2795.21 & -2795.21   \\
\bottomrule
\end{tabular}
    \caption{Number of training steps $\eta_{\varepsilon}$ needed for the likelihood $l^{(\eta)}_{K_0}(\hat{\theta}| \mathbf{y})$ to be within $\varepsilon$ of $l_{\textrm{min}}$ when using respective $K_0$,  across the synthetic case studies.  $\text{BIC}(\hat{K}_0^{(\eta_{\varepsilon}})$ corresponds to $l^{(\eta)}_{K_0}(\hat{\theta}| \mathbf{y})$. }
    \label{tab:conv_tol}
\end{table}

\section{Supplementary Tables and Figures}

\begin{table}[h!]
\begin{tabular}{lrrrrrrrrrrr}
\toprule
Range &  BIC &  BIC &  len &  addtv  &  non- &  num & num & num & rough  &  rough &  rough   \\
  & max & min &  & comps & stat. & age & year & coh & age & year & coh \\
   \midrule  \multicolumn{12}{c}{JPN Female ${\cal K}_r$: Rerun} \\ \midrule
 1-10 & -2723.53 & -2725.27 & 4.00 & 1.00 & 0\% & 1.00 & 1.50 & 1.50 & 0\% & 100\% & 100\% \\
 1-50 & -2720.11 & -2725.27 & 4.48 & 1.02 & 10\% & 1.12 & 1.82 & 1.54 & 0\% & 100\% & 100\% \\
 51-100 & -2717.87 & -2720.11 & 5.16 & 1.24 & 14\% & 1.28 & 2.30 & 1.58 & 0\% & 100\% & 100\% \\
\midrule  \multicolumn{12}{c}{JPN Female trained on ${\cal D}_{rob}$ and search in ${\cal K}_r$ } \\ \midrule
1-10 & -2716.89 & -2725.27 & 4.10 & 1.00 & 0\% & 1.00 & 1.70 & 1.40 & 0\% & 100\% & 100\% \\
 1-50 & -2713.77 & -2725.27 & 4.64 & 1.20 & 2\% & 1.20 & 1.88 & 1.56 & 0\% & 100\% & 100\% \\
 51-100 & -2712.33 & -2714.64 & 5.00 & 1.40 & 6\% & 1.22 & 1.88 & 1.90 & 0\% & 100\% & 100\% \\
\bottomrule
\end{tabular}
\caption{Additional GA runs on JPN Female HMD data \label{tab:JPN-Female-supplementary}}
\end{table}

\begin{figure}
    \centering
    \begin{tabular}{cccc}
    JPN Female & USA Male & JPN Male & SWE Female \\
    \includegraphics[width=0.24\textwidth,trim=0.2in 0.in 0.2in 0in]{Figures/resid_JPN_Female_complex.png} &
    \includegraphics[width=0.24\textwidth,trim=0.2in 0.in 0.2in 0in]{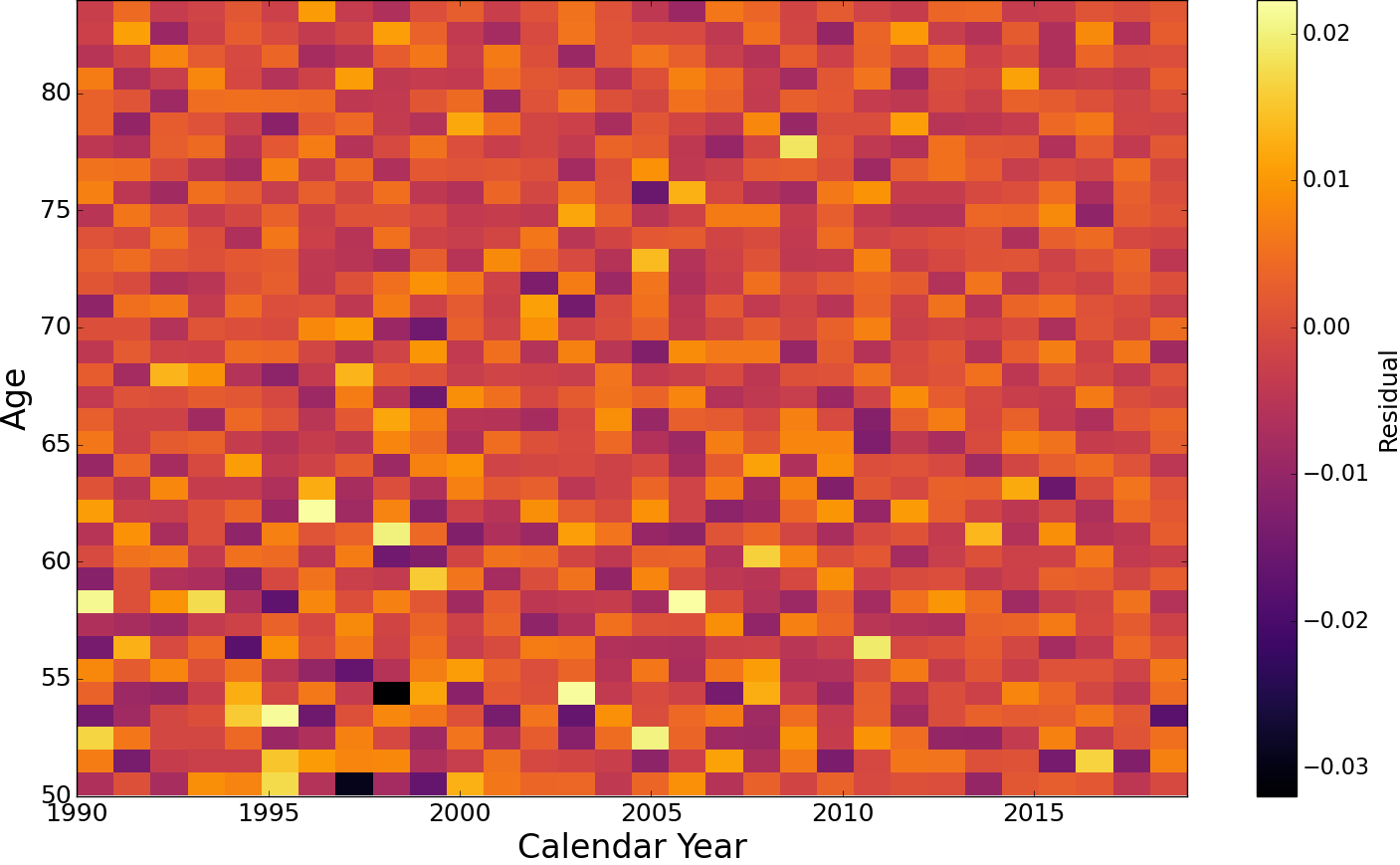} &
    \includegraphics[width=0.24\textwidth,trim=0.2in 0.in 0.2in 0in]{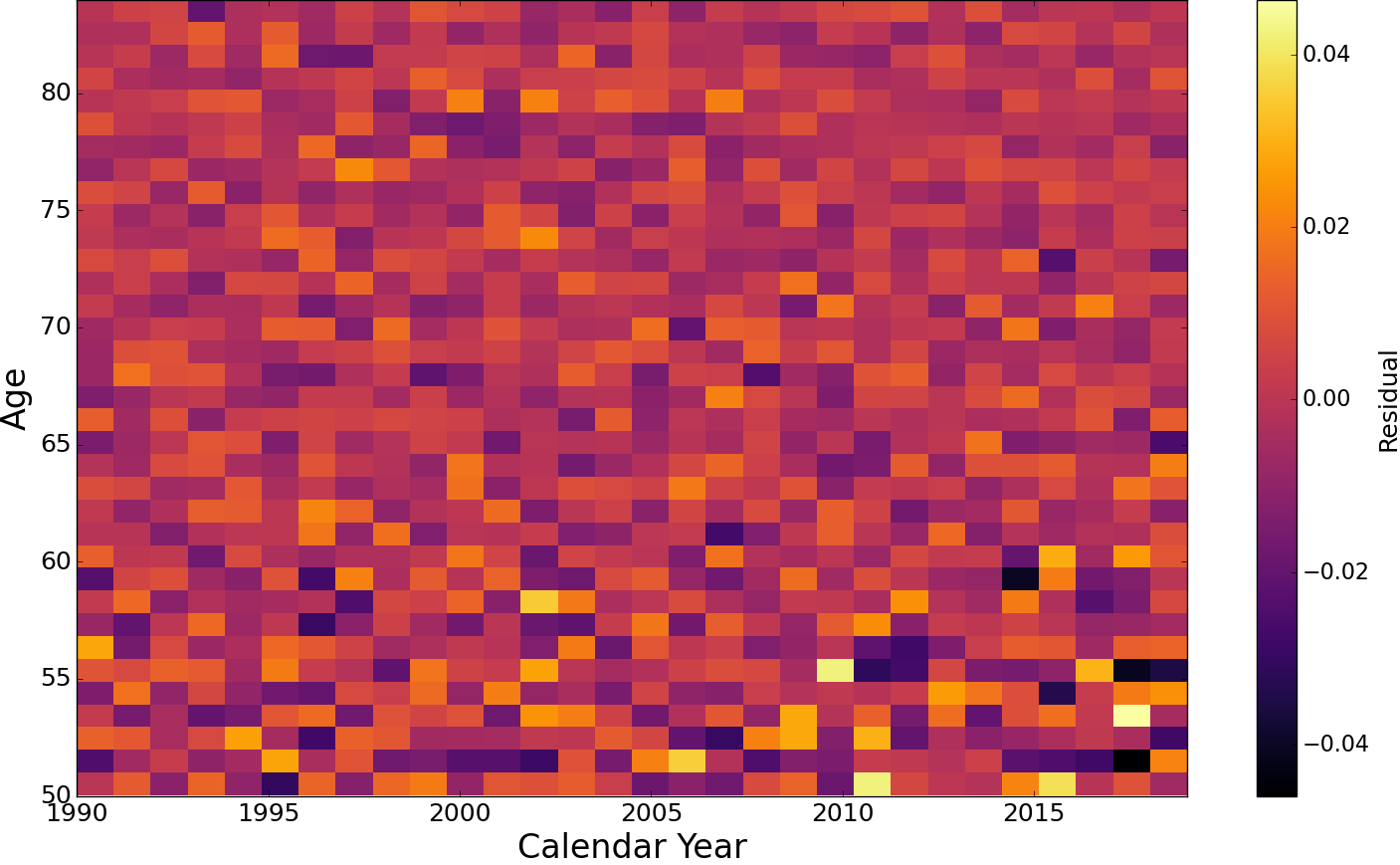} &
    \includegraphics[width=0.24\textwidth,trim=0.2in 0.in 0.2in 0in]{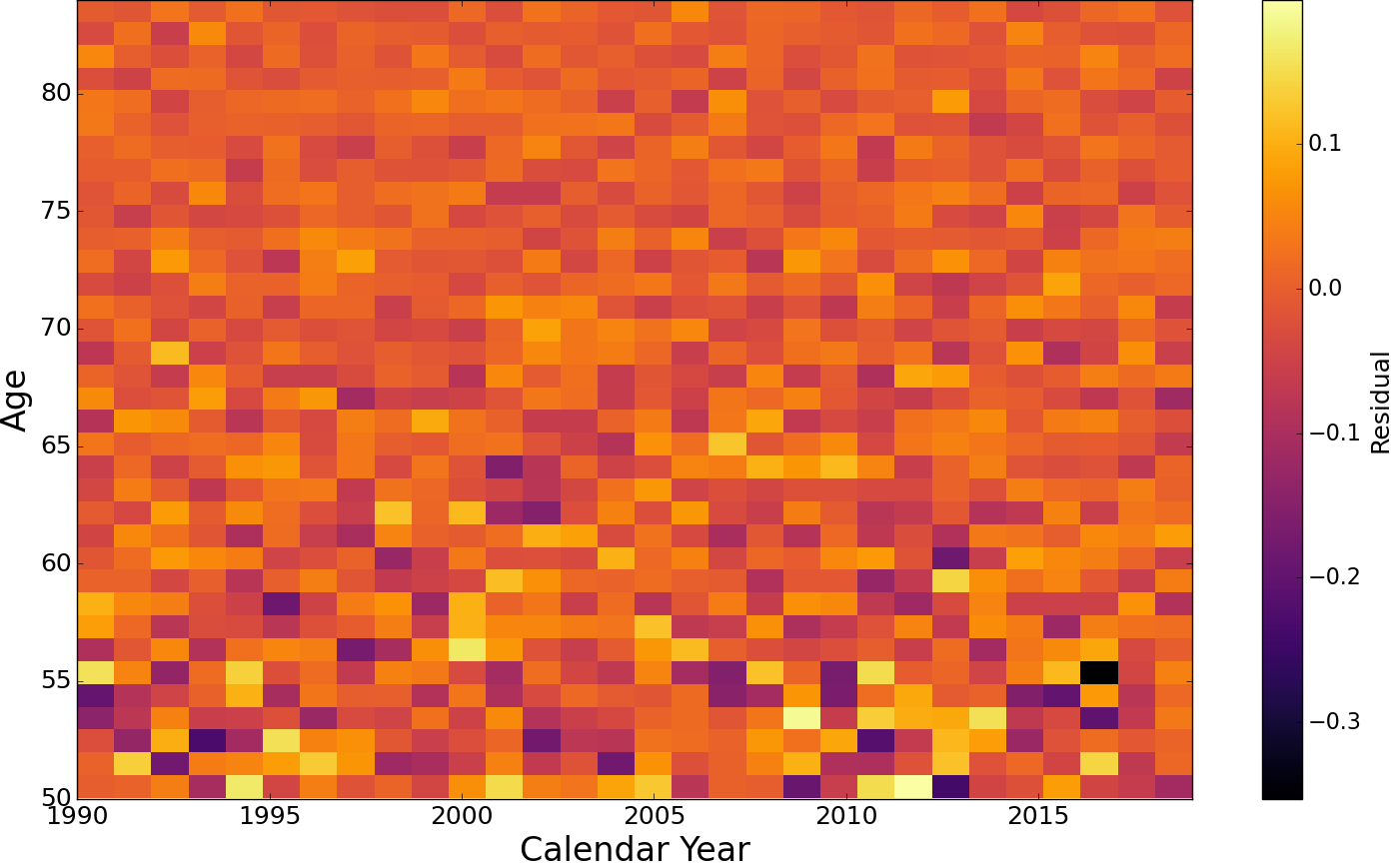} \\
      \includegraphics[width=0.24\textwidth,trim=0.2in 0.in 0.2in 0in]{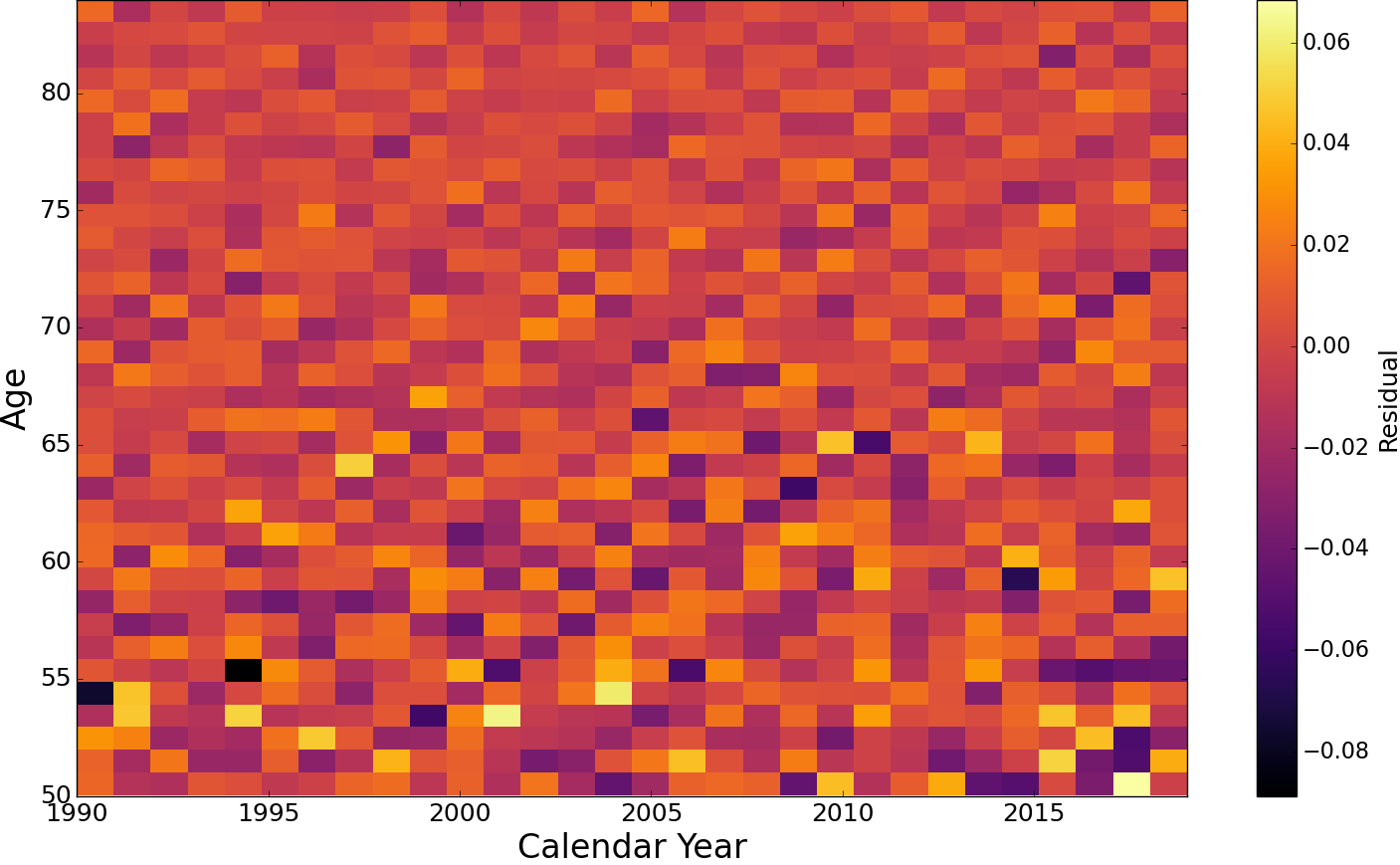} &
    \includegraphics[width=0.24\textwidth,trim=0.2in 0.in 0.2in 0in]{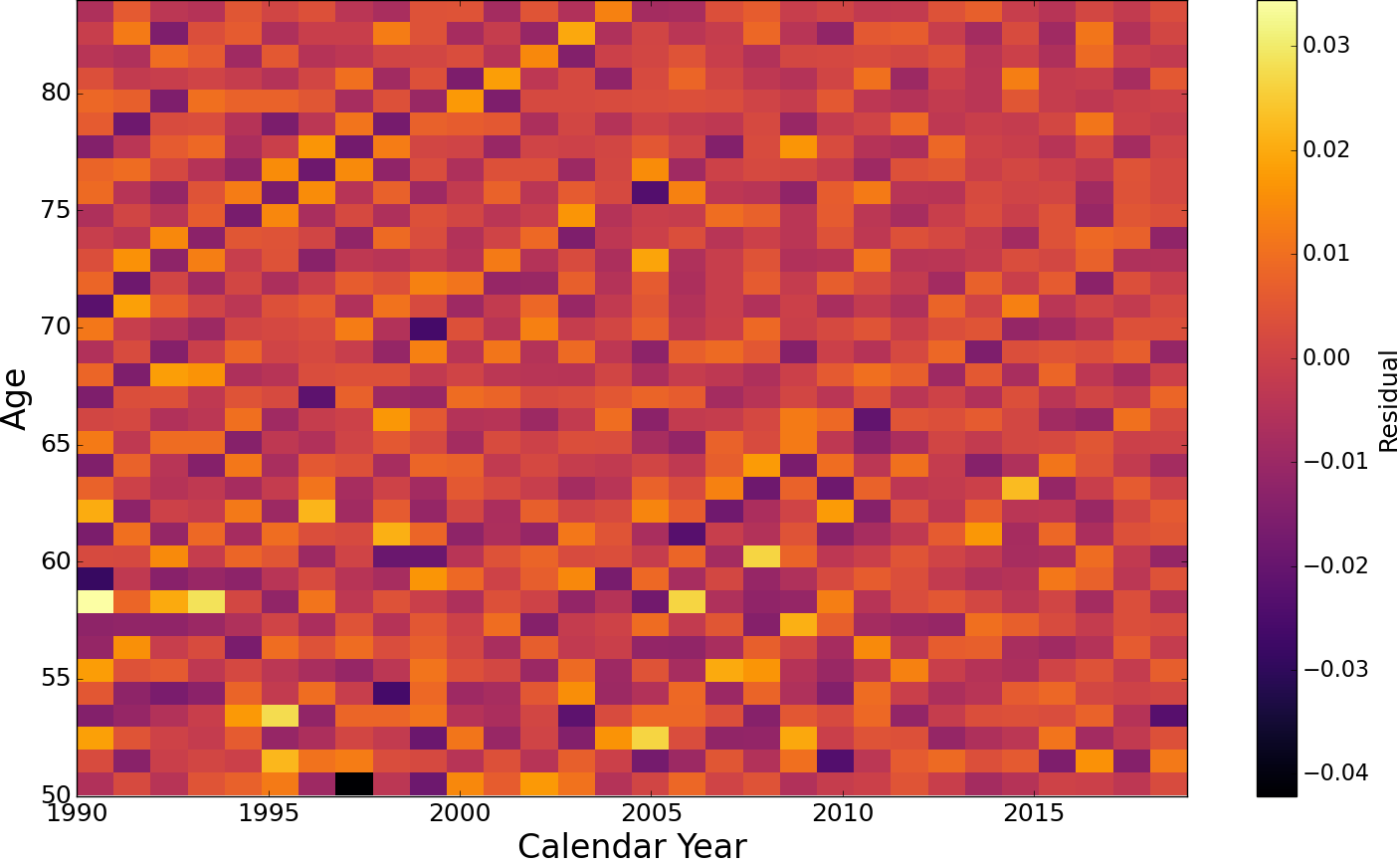} &
    \includegraphics[width=0.24\textwidth,trim=0.2in 0.in 0.2in 0in]{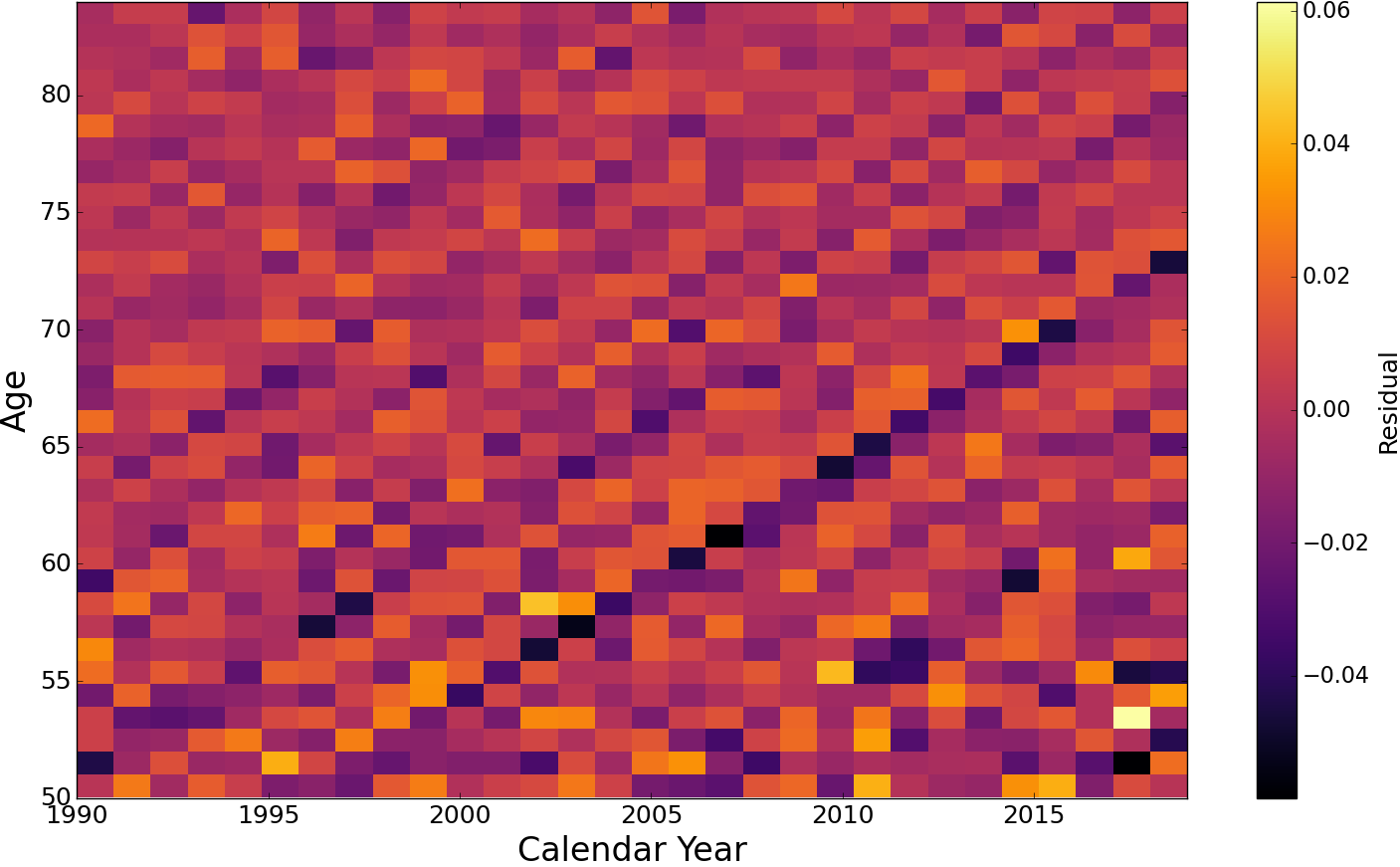} &
    \includegraphics[width=0.24\textwidth,trim=0.2in 0.in 0.2in 0in]{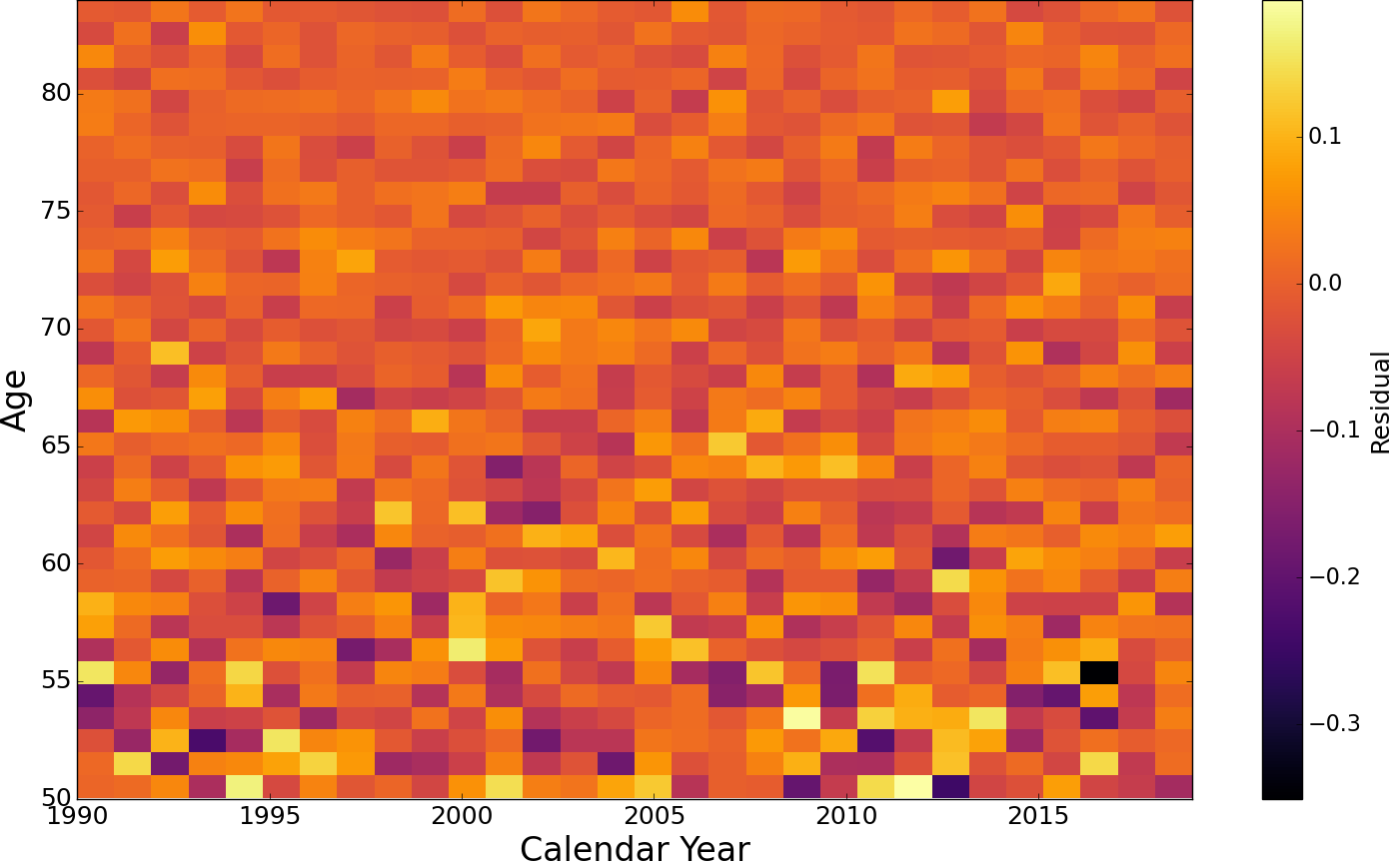} \\

    \end{tabular}
    \caption{Residual plots for the top kernel in ${\cal K}_f$ (top row) and the best kernel without a Cohort term (bottom row). The bottom row kernels are: JPN Female ($0.0144\cdot [0.0133\cdot \Meh_a(0.1)\cdot \AR2_y(1.45,0.26)+30.5479\cdot \M32_y(2.43)]\cdot \M52_a(1.12) \cdot \M12_y(62.46) $; JPN Male: $0.082 \cdot M52_a(0.91)\cdot \M52_y(0.9)\cdot [0.0114\cdot AR2_y(4.53,0.28)\cdot \RBF_a(0.13)+1.0985\cdot \M12_y(19.6)] $; USA Male: $0.0194 \cdot \Chy_a(0.78)\cdot \M12_a(22.69) \cdot \Chy_y(0.55) \cdot [1.7718\cdot \M12_y(25.09)+0.0111\cdot \M32_a(0.07)\cdot \RBF_y(0.09)]$ and SWE Female: $0.1125 \cdot \Chy_a(0.85) \cdot \Meh_y(1.16)\cdot \M12_y(37.74)$.}
    \label{fig:resid_nocohort}
\end{figure}

\section{Interpreting Bayes Factors}\label{sec:bf_appendix}

\begin{table}[t]
    \centering
    \begin{tabular}{lccl}
        $\text{BF}(K_1, K_2)$ & $\log\text{BF}(K_1, K_2)$ & Code & Interpretation \\ \midrule
        $> 100$ & $> 4.61$ & & Decisive evidence for $K_1$\\
        $30$--$100$ & $3.40$ to $4.61$ & . & Very strong evidence for $K_1$\\
        $10$--$30$ & $2.30$ to $3.40$ & * & Strong evidence for $K_1$\\
        $3$--$10$ & $1.10$ to $2.30$ & ** & Substantial evidence for $K_1$\\
        $1$--$3$ & \hspace{0.5cm}$0$ to $1.10$ & *** & Not worth more than a bare mention\\
    \end{tabular}
    \caption{Evidence categories for Bayes factors given by \cite{jeffreys1961theory}, assuming $l(K_1)<l(K_2)$.  Note that for the second column $\text{BIC}(K_2) - \text{BIC}(K_1) \approx \log\text{BF}(K_1, K_2)$.  Code column indicates a level of significance, where more $*$'s suggest less evidence that the GP distributions are different.}
    \label{tab:bf_evidence_categories}
\end{table}

\end{document}